\documentclass[runningheads]{llncs}

\usepackage{eccv}

\usepackage{eccvabbrv}

\usepackage{graphicx}
\usepackage{booktabs}

\usepackage[accsupp]{axessibility}  % Improves PDF readability for those with disabilities.
\usepackage{epsfig}
\usepackage{amsmath}

\usepackage{amssymb}
\usepackage{multirow,caption}
\usepackage{subcaption}
\usepackage{comment}
\usepackage{arydshln}
\usepackage{enumitem}
\usepackage[normalem]{ulem}
\usepackage{colortbl}
\usepackage{nicematrix}
\usepackage{tikz}
\usepackage{booktabs}
\usepackage{multirow}
\usepackage{lipsum}
\usepackage{cancel}
\usepackage{soul}
\usepackage{tabularx}
\usepackage{indentfirst}
\usepackage{algorithm}
\usepackage{algpseudocode}
\usepackage{amssymb}% http://ctan.org/pkg/amssymb
\usepackage{pifont}% http://ctan.org/pkg/pifont
\usepackage{hyperref}

\usepackage{orcidlink}

\makeatletter
\newcommand{\printfnsymbol}[1]{%
  \textsuperscript{\@fnsymbol{#1}}%
}
\renewcommand*{\@fnsymbol}[1]{\ifcase#1\or*\or$\dagger$\or\else\@arabic{#1}\fi}
\makeatother

\begin{document}
% \footnote{An example footnote.}
% ---------------------------------------------------------------
% TODO REVIEW: Replace with your title
\title{Versatile Incremental Learning: Towards Class and Domain-Agnostic Incremental Learning} 
% \footnote{An example footnote.}
% TODO REVIEW: If the paper title is too long for the running head, you can set
% an abbreviated paper title here. If not, comment out.
\titlerunning{Versatile Incremental Learning}

% TODO FINAL: Replace with your author list. 
% Include the authors' OCRID +++for the camera-ready version, if at all possible.
\author{
Min-Yeong Park\thanks{Equally contributed}\orcidlink{0009-0007-4143-7283} \and
Jae-Ho Lee\printfnsymbol{1}\orcidlink{0009-0003-0960-824X} \and
Gyeong-Moon Park\thanks{Corresponding author}\orcidlink{0000-0003-4011-9981}
}

% TODO FINAL: Replace with an abbreviated list of authors.
\authorrunning{M.-Y. Park et al.}
% First names are abbreviated in the running head.
% If there are more than two authors, 'et al.' is used.

% TODO FINAL: Replace with your institution list.
\institute{Kyung Hee University, Yongin, Republic of Korea \\
\email{\{pmy0792, jh.lee, gmpark\}@khu.ac.kr}} 

\maketitle
\begin{abstract}
% Modern continual learning methods have achieved notable progress in Class IL (CIL), and Domain IL (DIL) separately. However, in real scenarios that deep learning models encounter, we cannot assume how the distribution of input data will change, whether class increases or domain increases. Rather, it's reasonable to consider that the class or domain can be incremented at any time. In this context, we suggest a new setting, coined Versatile IL (VIL) where it is unknown for models whether class increases or domain increases in the subsequent task. The challenge of VIL is that the network cannot accumulate knowledge inherent in sequential tasks, unlike CIL and DIL. This problem causes severe shift in parameter space, exacerbating catastrophic forgetting. To solve this, we propose ...
Incremental Learning (IL) aims to accumulate knowledge from sequential input tasks while overcoming catastrophic forgetting.
% The mainstream scenarios within IL typically fall into two classes: Class IL (CIL) where tasks possess disjoint label spaces, and Domain IL (DIL) where tasks share the same label space but exhibit distinct distributions.
Existing IL methods typically assume that an incoming task has only increments of classes or domains, referred to as Class IL (CIL) or Domain IL (DIL), respectively.
In this work, we consider a more challenging and realistic but under-explored IL scenario, named \textit{Versatile Incremental Learning (VIL)}, in which a model has no prior of which of the classes or domains will increase in the next task.
% Existing methods encounter severe forgetting when incoming tasks change capriciously since they fail at accumulating knowledge in a proper direction and knowledge is over-written on the classifier continuously. 
% Existing methods mainly considered only \textit{inter-class} or \textit{inter-domain}, relying on a bias that the incoming task has only increments of classes or domains. However,
In the proposed VIL scenario, the model faces \textit{intra-class domain} confusion and \textit{inter-domain class} confusion, which makes the model fail to accumulate new knowledge without interference with learned knowledge.
To address these issues, we propose a simple yet effective IL framework, named \textit{Incremental Classifier with Adaptation Shift cONtrol (ICON)}.
Based on shifts of learnable modules, we design a novel regularization method called Cluster-based Adaptation Shift conTrol (CAST) to control the model to avoid confusion with the previously learned knowledge and thereby accumulate the new knowledge more effectively.
Moreover, we introduce an Incremental Classifier (IC) which expands its output nodes to address the overwriting issue from different domains corresponding to a single class while maintaining the previous knowledge.
% It expands, but finally squeezes the core knowledge and yields a worthy single output. 
We conducted extensive experiments on three benchmarks, showcasing the effectiveness of our method across all the scenarios, particularly in cases where the next task can be randomly altered. Our implementation code is available at \url{https://github.com/KHU-AGI/VIL}.

% that can be configured not only in traditional incremental scenarios but also in new ones: iDigits, CORe50, and DomainNet
 % 추후 더 추가.
\end{abstract}    

\section{Introduction}
\label{sec:intro}
\begin{figure*}[t!]
\centering
\begin{subfigure}[b]{0.323\columnwidth}
      \includegraphics[width=\textwidth]{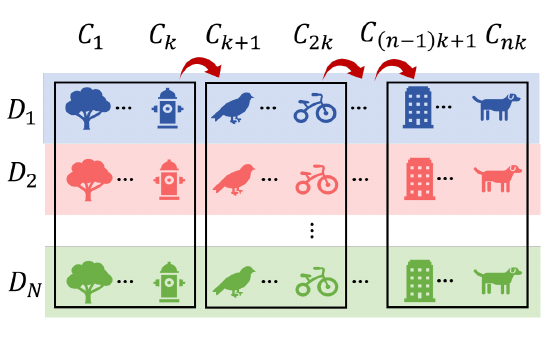}
      \caption{Class IL (CIL).}
      \label{fig:cil}
\end{subfigure}
\begin{subfigure}[b]{0.323\columnwidth}
      \includegraphics[width=\textwidth]{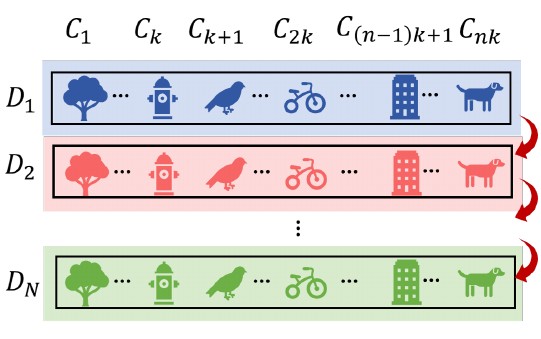}
      \caption{Domain IL (DIL).}
      \label{fig:dil}
\end{subfigure}
% \begin{subfigure}[b]{0.24\columnwidth}
%       \includegraphics[width=\textwidth]{figure/cdil.png}
%       \label{fig:cdil}
%       \caption{Cross-Domain IL (CDIL).}
% \end{subfigure}
% \hfill
\begin{subfigure}[b]{0.323\columnwidth}
      \includegraphics[width=\textwidth]{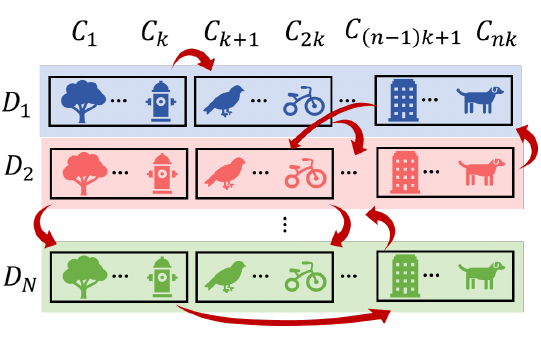}
      \caption{Versatile IL (VIL).}
      \label{fig:vil}
\end{subfigure}
\caption{Illustration of several IL scenarios, including our proposed new scenario, Versatile Incremental Learning. Each color shade indicates each domain group, and the solid box indicates each class group. The incremental step follows the red arrow.}
\label{fig:scenario}
\end{figure*}

% When deep learning models are to be deployed to the real world, it is likely that they encounter issues with distribution change of input data.  In order to learn incrementally from data of new distribution, it is important to minimize catastrophic forgetting, which infers to a phenomenon that a model fails at maintaining previously learned knowledge while dealing with new input stream.
\noindent Recently, Incremental Learning (IL) strategies \cite{rusu2016progressive,li2016learning,kirkpatrick2017overcoming,yoon2018lifelong,lee2017overcoming,friedemann2017continual,shin2017continual,yoon2018lifelong,li2019learn,hung2019compacting,yin2020dreaming,yan2021dynamically,smith2021always,douillard2022dytox,gao2022r,wang2022sprompts,wang2022learning,wang2022dualprompt,smith2023coda,gao2023unified,wang2023task,seo2023lfs,moon2023online,madaan2023heterogeneous,park2024pre} have made significant progress in leveraging deep neural networks in a situation when multiple input tasks arrive sequentially. The main challenge of IL is catastrophic forgetting \cite{mccloskey1989catastrophic}, which refers to a phenomenon in the model that significantly forgets what it has learned previously. The mainstream scenarios to tackle catastrophic forgetting within IL typically fall into two categories: Class IL (CIL) where tasks possess disjoint label spaces within the same domain (see Figure \ref{fig:cil}, and Domain IL (DIL) where tasks share the same label space but exhibit distinct distributions (see Figure \ref{fig:dil}. Most of the recent IL studies have focused on either CIL or DIL scenarios \cite{wang2022sprompts,wang2022learning,wang2022dualprompt,smith2023coda,gao2023unified}.
% of continual learning is divided into Class Incremental Learning (CIL) and Domain Incremental Learning (DIL). CIL is learning new classes for each task within the same domain. On the other hand, models for DIL learn the same classes from different domains sequentially for each task. 

% \begin{figure}[t!]
% \begin{subfigure}[b]{0.495\columnwidth}
%       \includegraphics[width=\textwidth]{figure/cil.png}
%       \label{fig:cil}
% \end{subfigure}
% \hfill
% \begin{subfigure}[b]{0.495\columnwidth}
%       \includegraphics[width=\textwidth]{figure/dil.png}
%       \label{fig:dil}
% \end{subfigure}
% \newline
% \begin{subfigure}[b]{0.495\columnwidth}
%       \includegraphics[width=\textwidth]{figure/cdil.png}
%       \label{fig:cdil}
% \end{subfigure}
% \hfill
% \begin{subfigure}[b]{0.495\columnwidth}
%       \includegraphics[width=\textwidth]{figure/vil.png}
%       \label{fig:vil}
% \end{subfigure}
% \vspace{-10mm}
% \label{fig:scenario}
% \caption{Illustration of all kind of continual learning scenarios. Our proposed new Versatile Incremental Learning (d) also . The shade by color means each domain group, and the solid box means each class group, and the incremental step proceeds according to the arrow.}
% \end{figure}

% The majority of recent continual learning works \cite{wang2022sprompts,wang2022learning,wang2022dualprompt,smith2023coda,gao2023unified} have focused on only CIL or DIL.
% dealing with scenarios where the input data has the same domain but an increasing number of classes, or where the input data has the same number of classes but an increasing number of domains. 
These existing settings are based on a strong assumption that sequential input tasks always share the same classes or domains, \ie, only classes or domains can increase, and this assumption makes the existing methods impractical to apply to the real world. For example, the models for self-driving cars should continuously learn increasing classes of objects while the domains where a car lies continuously change by different environments (e.g., weather conditions, regions, etc.).
% environment of the car lies can be moved to another region. 
% Therefore, it is important that the networks are able to deal with unexpected input tasks not knowing how the task increases. 
Therefore, the models need to learn new classes or domains sequentially when they cannot expect what will increment afterward.
Yet, this situation is under-explored in the existing IL settings although it is crucial for the model function well in real-world scenarios. 

% This work is the first approach to formulate the situation where the network is unaware of what increases in a consecutive task stream, and propose effective methods to tackle with challenges in the situation. We clarify the challenges in the scenario with our empirical finding that existing CIL or DIL methods \cite{wang2022sprompts,wang2022learning,wang2022dualprompt,smith2023coda,gao2023unified} are shown to fail when the next input task changes suddenly. 
% Existing CIL or DIL models \cite{wang2022sprompts,wang2022learning,wang2022dualprompt,smith2023coda,gao2023unified} are shown to fail when the next input task changes suddenly. 
In this paper, to alleviate the aforementioned assumption, we introduce a new IL scenario called \textbf{Versatile Incremental Learning} (\textbf{VIL}) for the first time, which is more challenging and realistic than the existing CIL or DIL settings. VIL aims to deal with a situation where the incoming tasks can contain new classes in the same domain, the same classes in a new domain, or new classes in a new domain. In the VIL setting, the model encounters new tasks without knowing how these tasks will increase, as depicted in Figure \ref{fig:vil}. In this class and domain-agnostic incremental scenario, the goal is for the models to learn how to accumulate task-specific knowledge continuously without forgetting, regardless of the incremental type of incoming tasks.
% Networks successfully trained for VIL can be applied to scenarios when there is no guarantee of which new tasks will increase, which is more practical than CIL and DIL.
\begin{figure}[!t]
\centering
\includegraphics[width=\textwidth]{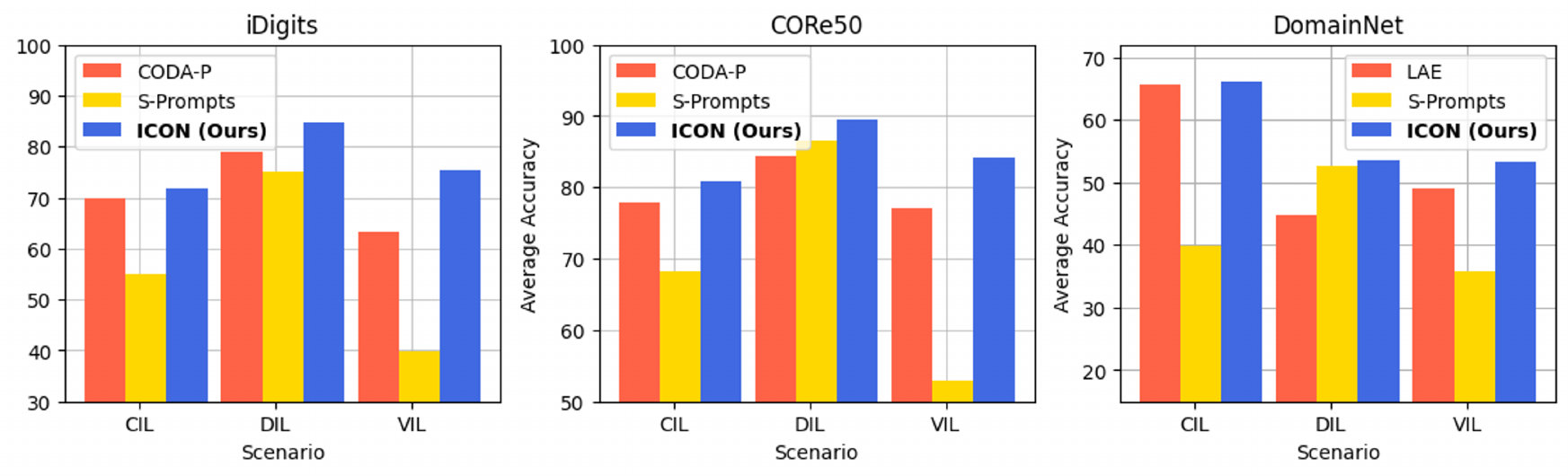}
\label{fig:motivation1}
\caption{Comparison of average accuracies among existing CIL and DIL methods in iDigits, CORe50, and DomainNet. In this figure, we compare the baselines that show the best performances in each benchmark, e.g., CODA-Prompt \cite{smith2023coda} and S-Prompts \cite{wang2022sprompts} in iDigits and CORe50, and LAE \cite{gao2023unified} and S-Prompts \cite{wang2022sprompts} in DomainNet. Our proposed ICON outperforms the previous state-of-the-art methods in all scenarios, including the challenging VIL setting.
}
\label{fig:motivation}
\end{figure}

To investigate how the novel VIL scenario is challenging, we conducted experiments on three different datasets. As shown in Figure \ref{fig:motivation}, existing CIL and DIL methods fail in the VIL scenario.
% Since CIL methods do not consider the possibility of increasing input domains, they are ineffective when domains are incremented, and vice versa for DIL methods.
% For the same reason, DIL methods cannot adapt when new classes increase. 
% To tackle the challenges of this crucial scenario, we propose a new setting and the solutions for the challenges in the new setting.
We analyze that existing CIL methods face severe drift in the classifier while learning new domains that share the same classes.
% Specifically, the classifiers of existing CIL methods are severely drifted after learning new domains for the same classes. 
% Even though they can learn new tasks via adjustable modules (prompts, adapters, \etc), the overwriting of knowledge from new domains in classifiers intensifies catastrophic forgetting in VIL. 
% In the case of DIL methods, they fail on VIL due to their inability to select task-specific modules at the inference time. The utility of task-specific modules heavily relies on the accuracy of their selection, which can severely harm the robustness of DIL methods in the VIL setting.
% In the case of DIL methods, they fail on VIL since they only consider learning fixed classes and cannot deal with new class distribution effectively.
In the case of DIL methods, they fail on VIL due to inter-domain class confusion, which is caused by their strong assumption of increasing only the domains.
% \my{Therefore, preventing severe drift in the classifier and adapting to distribution changes by increased classes in a same domain are the key challenges in the VIL scenario.
Therefore, VIL is a challenging scenario that causes severe catastrophic forgetting due to drift in the classifier and class distribution change.
% Hence, it is important and at the same time challenging to learn capricious input tasks effectively.

In light of these empirical findings, we propose a new method named \textbf{I}ncre-mental \textbf{C}lassifier with Adaptation Shift c\textbf{ON}trol (\textbf{ICON}), which tackles the VIL setting to consider not just one kind of CIL or DIL, but both when learning incrementally. The main challenge of VIL is that since subsequent tasks can have random incremental types, it is difficult for the model to accumulate knowledge in a specific direction for each type of incremental task. To this end, we propose a simple yet effective strategy coined \textit{Cluster-based Adaptation Shift conTrol (CAST)} to control the learning direction of the model in a stream of erratic input tasks. Specifically, we effectively regularize the direction of current learning concerning the learning directions of previous tasks. Furthermore, we propose \textit{Incremental Classifier (IC)}, a new strategy to regulate the classifier by increasing its output nodes dynamically. It helps the model to learn knowledge of multiple domains for each class effectively while preventing severe forgetting in the classifier.
% to learn already seen classes which lie in domains with severe gap. 
Through these strategies, our proposed ICON achieves state-of-the-art performances in the VIL scenario, including existing IL scenarios across three benchmarks. Our main contributions can be summarized as follows:
\begin{itemize}
    \item We propose a new realistic IL scenario for the first time, coined Versatile Incremental Learning (VIL), where a model has no prior knowledge of how sequential tasks possess class or domain distributions. To tackle this challenging scenario effectively, we propose a new IL framework, called Incremental Classifier with Adaptation Shift cONtrol (ICON).
    % This new problem setting presents a challenge for learning agents to effectively learn new tasks, regardless of whether the class or domain of input data increases.
    \item We introduce a new Cluster-based Adaptation Shift conTrol (CAST) loss to guide the learning direction of subsequent tasks to avoid colliding with those of dissimilar tasks already learned.   
    \item To effectively learn multiple domains in the VIL scenario, we propose a new Incremental Classifier (IC) that dynamically increases the output units corresponding to a single label. This approach can alleviate severe drift in the output nodes of the classifier.
    \item Comprehensive experiments demonstrate that the proposed ICON outperforms the existing state-of-the-art methods significantly in the VIL setting, as well as existing IL scenarios, which shows the effectiveness of the proposed framework.
    % ICON effectively regularizes the direction of current learning and prevents the over-writing problem of output nodes. 
\end{itemize}

\section{Related Work}\label{sec:relatedwork}

% \begin{figure}[t!]
%     \begin{subfigure}[b]{0.495\columnwidth}
%           \includegraphics[width=\textwidth]{figure/DDDC.png}
%           \label{fig:dddc}
%     \end{subfigure}
%     \begin{subfigure}[b]{0.495\columnwidth}
%       \includegraphics[width=\textwidth]{figure/CCCD.png}
%       \label{fig:cccd}
%     \end{subfigure}
%     \vspace{-10mm}

% \caption{}
% \label{fig:shifts}
% \end{figure}

\noindent\textbf{More Realistic IL Scenario.} Recently, cross-domain IL, which sequentially learns classes from different domains, has begun to be studied \cite{buzzega2020dark, xie2022general, simon2022generalizing, wang2023isolation}. This is a more difficult and realistic scenario than traditional class IL or domain IL, because the large domain gap between each task, coupled with the learning of discrete classes for each task, poses challenges for knowledge transfer.
% However, in an offline manner, this setting is forced to use only a fraction of the dataset because of the difference in the number of domains and classes that can be divided discretely (incremental steps), and these unfair results in significant differences depending on how it is configured \cite{buzzega2020dark,xie2022general, wang2023isolation}. 
% Moreover, from a prior knowledge perspective, the prior that classes increase for each task is the same as traditional incremental learning, so there is a considerable disparity from reality still. 
However, the existing methods \cite{buzzega2020dark, simon2022generalizing, wang2023isolation} restrict this setting in which each subsequent task always has unseen classes on different domains. Another work \cite{xie2022general} considers the setting that early tasks have only increasing classes, and the others have only increasing domains. This setting does not consider that classes or domains can increase at any time in the real world.
In this paper, we introduce a more realistic and general scenario where the model learns consecutive tasks without prior knowledge of how inputs will increase. The absence of prior knowledge regarding the increment type allows the VIL scenario to encompass not only traditional IL scenarios (CIL and DIL) but also more realistic cross-domain IL as a subset of task streams.

% and propose an incremental classifier and a regularization method that enable learn effectively to a various input types.
\begin{figure}[t!]
\centering
\includegraphics[width=0.8\textwidth]{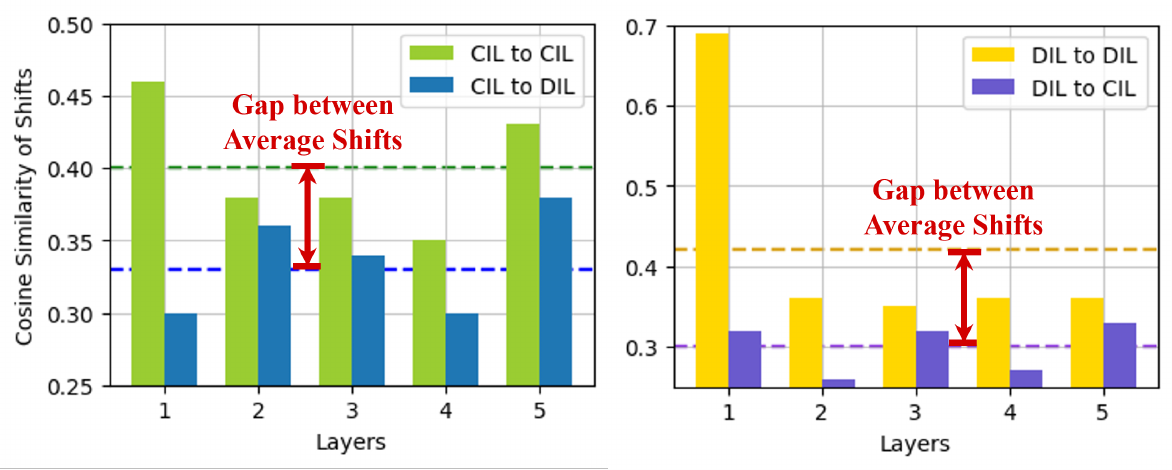}
\label{fig:shifts}
\caption{Illustration of comparison of shifts in adapters when the type of IL remains the same or changes in DomainNet. Shifts are measured by subtracting the previous weights with weights after learning a task.}
% \caption{Illustration of comparison of gradients of each layer in adapters when the type of incremental learning remains the same or changes in iDigits. Gradients are accumulated for each task, then the Kullback-Leibler divergence is measured to compare the shift in the direction of gradients.}
\label{fig:shifts}
\end{figure}

\noindent\textbf{Regularization for IL.} Regularization methods for IL have evolved to prevent catastrophic forgetting by adding regularization terms with reference to the old model. In general, the existing regularization methods can be categorized into weight regularization and function regularization \cite{wang2023acomprehensive}. EWC \cite{kirkpatrick2017overcoming}, SI \cite{friedemann2017continual}, and MAS \cite{rahaf2018memory} are the classic regularization approaches that calculate the importance of parameters and consider it to avoid severe changes of important parameters by imposing a penalty to the loss function. Function regularization methods maintain the original knowledge using the output of the model. Using only current task data, LwF \cite{li2016learning} distills the knowledge of the previous model by matching its outputs with the current ones. 
% The previous approaches to regularize the learning were effective, but since they impose a fixed form of regularization regardless of the stream of input sequence, they are less effective when the input stream is changing capriciously. 

However, existing regularization methods are not optimal when the tasks increase in a versatile way. We empirically investigated how the parameters are updated when the type of IL changes compared to when the type of IL remains the same.
As shown in Figure \ref{fig:shifts}, the similarities of shifts of the model between the same type of IL are fairly higher than those when what increments in the following task changes. This indicates that the shifts of the model lean towards disparate directions when the type IL shifts from one to another. Therefore, when its learning direction shifts, the model cannot accumulate knowledge well, deteriorating catastrophic forgetting. Hence, it is needed to regularize the direction of learning of the model not to conflict with the direction of dissimilar learning in history. Inspired by these points, we propose a novel regularization method for this sake, which will be further described in Section \ref{sec:3.2}.

\noindent\textbf{Model Expansion for IL.}
% To tackle the proposed scenario, we focus on designing dynamic expandable architecture.
Many attempts \cite{hung2019compacting, lee2017overcoming,yoon2018lifelong,li2019learn,yan2021dynamically,wang2023isolation,douillard2022dytox,hu2023dense,wang2023task} have been made to dynamically expand the models, \eg. neurons, modules, \etc., and the extended capacity is utilized to acquire task-specific knowledge in IL. DER \cite{yan2021dynamically} increases a new feature extractor per each task to learn task-specific super features with mask layers. Moreover, DyTox \cite{douillard2022dytox} proposed a dynamically expandable representation model as well as a classifier in a task-dynamic manner. Recently, ESN \cite{wang2023isolation} also proposes the expandable classifier method, but they leverage training samples augmented in different ways in the inference phase to vote for the classifier with low free energy (likely in-distribution) \cite{tang2021codes, liu2022model} among multiple classifiers. However, the aforementioned expandable classifiers also fail in our novel VIL setting because there is no consideration for the same class in different domains which causes \textit{intra-class domain} confusion and \textit{inter-domain class} confusion by weight overwriting. In contrast, we propose a novel selective node-wise expandable classifier to address the overwriting problem that leads to catastrophic forgetting in the existing classifier when learning the same classes across different domains.

\section{Method}
\label{sec:method}
In this section, we propose a simple yet effective incremental learning framework named ICON to address the problems of VIL aforementioned. We first introduce the proposed VIL scenario and briefly review the problems that derive from the under-explored VIL scenario in Section \ref{sec:3.1}. Next, we propose two novel methods: Cluster-based Adaptation Shift Control (CAST) and Incremental Classifier (IC) in Section \ref{sec:3.2} and Section \ref{sec:3.3}, respectively. Finally, we describe the whole training scheme with an optimization objective in Section \ref{sec:3.4}.

\subsection{Scenario Description of VIL}
\label{sec:3.1}
\noindent Previously, the types of IL can be categorized into CIL and DIL, which have limitations in that they assume that what will increase in the next task compared to the current task is fixed with class or domain. However, in the real-world, class, domain, or both can be increased at a time. Therefore, we introduce a new IL scenario named Versatile-Incremental Learning (VIL) which better suits for the real-world. In the VIL scenario, only classes, only domain or both can increase in the very following task and we describe each scenario in Table \ref{tab:scenario} and illustrate it in Figure \ref{fig:scenario}. As shown in Table \ref{tab:scenario}, the existing CIL setting is fixed to have only disjoint label space, and the DIL setting is fixed to have only disjoint domain space. However, in a task stream of VIL, two sequential tasks can have disjoint label space, disjoint domain space, or both. In these differences, the model in the VIL scenario suffers from the problems as follows.

In the proposed VIL scenario, the model fails on VIL due to intra-class domain confusion and inter-domain class confusion that is derived from the ever-dynamically changing input distribution. Furthermore, the model faces severe drift in the classifier while learning new domains that share the same classes. To address these problems, we propose Cluster-based Adaptation Shift conTrol (CAST) to prevent model confusion and Incremental Classifier (IC) to prevent weight drift from different domains corresponding to a single class. We precisely explain each proposed method in the following sections.

\newcommand{\xmark}{\ding{55}}%
\begin{table}[!t]
\caption{
Comparison of typical incremental learning scenarios and proposed VIL. 
}
\centering
\renewcommand{\arraystretch}{1.2}
\resizebox{0.95\textwidth}{!}{
{\scriptsize
\begin{tabular}{@{\hspace{1.5mm}}l@{\hspace{3mm}}@{\hspace{3mm}}c@{\hspace{5mm}}c@{\hspace{3mm}}c@{\hspace{3mm}}}
\specialrule{.1em}{.1em}{.1em} 
Scenario Attribute& CIL~&~DIL~&~\textbf{VIL} \\ 
\specialrule{.05em}{.1em}{.1em} 
Can it have disjoint label space? & \ding{51} & \xmark & \ding{52} \\ 
Can it have disjoint domain space?& \xmark & \ding{51} & \ding{52} \\ 
Can it have both disjoint label and domain space?& \xmark & \xmark & \ding{52} \\
\specialrule{.1em}{.1em}{.1em} 
\end{tabular}}}
\label{tab:scenario}
\end{table}

% \newcommand{\xmark}{\ding{55}}%
% \begin{table}[!t]
% \caption{
% Comparison of typical incremental learning scenarios and proposed VIL.
% \centering
% \renewcommand{\arraystretch}{1.2}
% \resizebox{0.95\textwidth}{!}{
% {\footnotesize
% \begin{tabular}{@{\hspace{1.5mm}}l@{\hspace{3mm}}@{\hspace{3mm}}c@{\hspace{5mm}}c@{\hspace{3mm}}}
% \specialrule{.1em}{.1em}{.1em} 
% Scenario& Can it have disjoint label space?&Can it have disjoint domain space?&Can it have both disjoint label and domain space? \\ 
% \specialrule{.05em}{.1em}{.1em} 
% CIL& \ding{51} & \xmark \\ 
% DIL& \xmark & \ding{51} \\ 
% \textbf{VIL}& \ding{51} or \xmark & \ding{51} or \xmark \\
% \specialrule{.1em}{.1em}{.1em} 
% \end{tabular}}}
% \label{tab:scenario}
% \end{table}

\begin{figure*}[!t]
\centering
\begin{subfigure}[b]{\textwidth}
      \includegraphics[width=\textwidth]{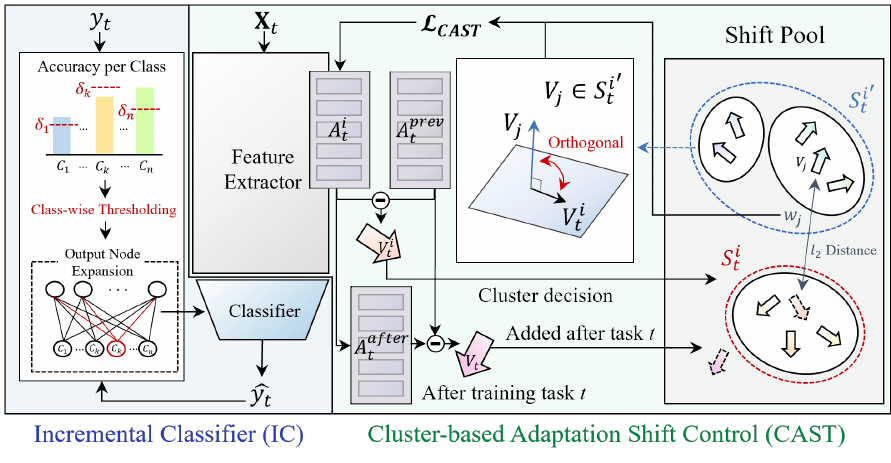}
      \label{fig:architecture}
\end{subfigure}
\caption{Architecture overview. In training time, the model calculates the current shift ${V_t^i}$ of learnable modules by subtracting them with previous ones. Then a cluster ${S_t^i}$ which ${V_t^i}$ belongs to is decided, and shifts in the the shift pool which belong to other clusters ${S_{t}^{i'}}$ are considered to be from disparate previous tasks. To guide the current learning toward a direction where it does not conflict with ${V_{j}}$, ${V_t^i}$ is regularized to be orthogonal to for all ${V_{j}}$ in ${S_{t}^{i'}}$. After learning a task, $V_t$ is saved as a shift in the shift pool which will be used for clustering afterwards.}
\label{fig:architecture}
\end{figure*}

\subsection{Cluster-based Adaptation Shift Control}\label{sec:3.2}
\noindent In existing CIL and DIL scenarios, the model can accumulate knowledge steadily without any guidance about updating directions of learnable weights since domains or classes are shared for the entire tasks, which makes the accumulation of knowledge easier. However, in the VIL scenario, there is no prior knowledge of what will increment and what will remain stationary in the following task whether it is classes or domain. Consequently, this makes it difficult to accumulate knowledge in a steady direction along the entire tasks. Therefore, it is necessary to have guidance for a model about how to accumulate the knowledge preventing its current shift from moving capriciously. Here, we propose \textbf{C}luster-based \textbf{A}daptation \textbf{S}hift con\textbf{T}rol (\textbf{CAST}) loss for this sake. 

In order to learn the current task without affecting the various knowledge learned previously, we regularize the direction of updates in current adapters with respect to the directions of updates in previous tasks. For this sake, the model saves weights of adapters ${A}_{t-1}^{prev}$ before learning task ${t-1}$ as shown in Figure \ref{fig:architecture}. 
After learning task ${t-1}$, ${V}_{t-1}$ which we define as the shift in adapter while learning the task is measured by subtracting ${A_{t-1}^{prev}}$ from ${A}_{t-1}^{after}$, where ${A}_{t-1}^{after}$ is the adapter weights after learning task $t-1$ as follows:
\begin{equation}\label{eq:1}
    \begin{aligned}
    V_{t-1} = A_{t-1}^{after} - A_{t-1}^{prev}.
    \end{aligned}
\end{equation}
Here, the subscripts of \textit{A} and \textit{V} indicate the task identity, while the superscripts of \textit{A} indicate the status with regard to the current task. The shift $V_{t-1}$ in the direction of task $t-1$ is then saved to a shift pool. As shown in Figure \ref{fig:architecture}, the shift pool saves all previous shifts obtained after learning each task.
It is followed by clustering the entire shifts in the shift pool saved until task ${t-1}$ using the $K$-Means algorithm. Then, when training the following task ${t}$, the shift of adapter for current iteration $i$, $V_{t}^{i}$ is calculated using the current weights of adapter $A_{t}^{i}$ for each iteration. Here, we define the shift in the direction of current learning in comparison with the state before the beginning of the task $t$ as follows:
\begin{equation}\label{eq:1}
    \begin{aligned}
    V_{t}^{i} = A_{t}^{i} - A_{t}^{prev},
    \end{aligned}
\end{equation}
where $A_{t}^{prev}$ is the adapter weights before learning task $t$. 

The subtraction of weights implies the meaning of the direction of learning the task, and it can be derived from the formula of classic gradient descent for parameter update as follows:
\begin{equation}\label{eq:3}
    \begin{aligned}
    A_{t}^{i+1} = A_{t}^{i} - \eta \frac{\partial \mathcal{L}}{\partial A_{t}^{i}},\quad 
    A_{t}^{1} = A_{t}^{0} - \eta \frac{\partial \mathcal{L}}{\partial A_{t}^{0}}.
    % \approx V_{t}^{i} \:.
    \end{aligned}
\end{equation}
\text{From Equation \ref{eq:3},}
\begin{equation}\label{eq:4}
    \begin{aligned}
    % \text{From Equation \ref{eq:3},} 
    \quad A_{t}^{i} = A_{t}^{0} - \eta \sum_{k=0}^{i-1}\frac{\partial \mathcal{L}}{\partial A_{t}^{k}},
    % \approx V_{t}^{i} \:.
    \end{aligned}
\end{equation}
\begin{equation}\label{eq:4}
    \begin{aligned}
    \therefore ~ V_t^i = A_{t}^{i} - A_{t}^{prev} = A_{t}^{i} - A_{t}^{0} = - \eta \sum_{k=0}^{i-1}\frac{\partial \mathcal{L}}{\partial A_{t}^{k}}.
    % \approx V_{t}^{i} \:.
    \end{aligned}
\end{equation}
We can replace $A_t^{i}$ using $A_t^{i-1}$, and after successive replacement from $A_t^i$ to $A_t^1$, the subtraction of two weights is expressed in the form of summation of gradients, which is accumulated gradients. Therefore, by simply subtracting two weights, we can utilize the shift in the direction of the current iteration with regard to the state before learning task $t$.

After the calculation of $V_{t}^{i}$, the model predicts the index of the cluster which $V_{t}^{i}$ belongs to for each iteration among clusters established via $K$-Means before learning task $t$. The prediction is done by selecting a cluster whose center is the closest with $V_{t}^{i}$. The cluster that ${V_{t}^{i}}$ belongs to is notated as ${S_{t}^i}$, and other clusters as $S_{t}^{i'}$. The shifts that belong to the rest of clusters $S_{t}^{i'}$ represent the directions of previous tasks whose directions are distinctive from current learning. Therefore, to prevent the direction of current learning ${V_{t}^{i}}$ from colliding with those directions, they are used to regularize the current learning. The regularization of the direction of current learning is done by making $V_{t}^{i}$ to be orthogonal with shifts in $S_{t}^{i'}$.
Finally, the equation for CAST loss is defined as follows:

\begin{equation}\label{eq:cast}
    \begin{aligned}
        \begin{gathered}
        \mathcal{L}_{CAST} =  \sum_{j} w_{j} \cdot \frac{V_{t}^i\cdot V_{j}}{\left \| V_{t}^i \right \|  \left \| V_{j}\right \|}, \quad w_{j} = \frac{\left \| V_{t}^i - V_{j}\right \|_{2}}{\sum_{V_k \in S_t^{i'}} \left \| V_{t}^i - V_{k}\right \|_{2}},  \\
        % \text{where }V_{j}\in S_{t}^{i'}, 
        % % \: S_{t}^{i'} \subset \{V_1, V_2, \dots , V_{t-1}\},
        % \: S_{t}^{i'} = \{ V_1, V_2, \dots, V_{t-1}\} - S_t^i,
        % \: \:w_{j}= \left \| V_{t}^i - V_{j}\right \|_{2}.
        \end{gathered}
    \end{aligned}
\end{equation}
where $V_{j}\in S_{t}^{i'}, 
        % \: S_{t}^{i'} \subset \{V_1, V_2, \dots , V_{t-1}\},
        \: S_{t}^{i'} = \{ V_1, V_2, \dots, V_{t-1}\} - S_t^i.
        % , \text{ and}
        % \: \:w_{j}= \left \| V_{t}^i - V_{j}\right \|_{2}$.
        $
We consider all shifts in $S_{t}^{i'}$ using weighted sum in the loss, with ${w_j}$ being acquired using Euclidean distance between ${V_t^i}$ and $V_{j}$ in $S_{t}^{i'}$, thereby considering the discrepancy of current shift and each $V_j$ differentially. 
As a result, the CAST loss leads the current shift not to affect the shifts in the shift pool while adapting to the current task. Specifically, the updates of weights in the current task are adjusted in the direction that preserves the direction of disparate tasks, while fine-tuning. 
Hence, the model can accumulate knowledge in a stable direction with regard to all tasks, even when the input tasks change arbitrarily.
As the sequence of tasks increases, the model can further benefit from CAST by regularizing the direction of shifts and thereby accumulating knowledge in succession.

\algnewcommand{\LineComment}[1]{\State \(\triangleright\) #1}
\renewcommand{\algorithmicrequire}{\textbf{Input:}}

\begin{figure}[t!]
\includegraphics[width=\textwidth]{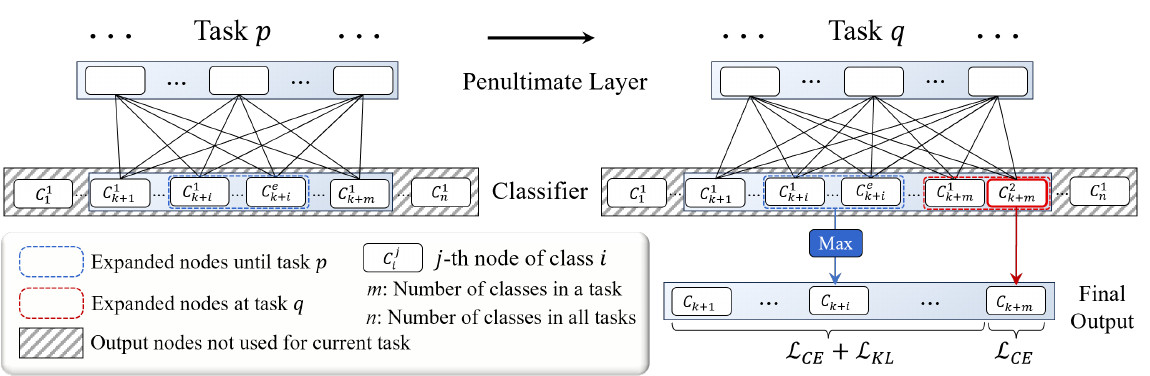}
\caption{Illustration of Incremental Classifier (IC) in training. The model increases the output node of classifier if needed whenever classes in current task $q$ had already learned before, \ie task $p$. Nodes for remaining classes included in task $q$ are trained to preserve the knowledge via distillation. The original nodes with classes whose output nodes has been increased at task $q$ are kept intact by omitting them from cross-entropy loss.}
\label{fig:ic}
\end{figure}

\subsection{Incremental Classifier}\label{sec:3.3} 
\noindent In the proposed VIL scenario, the existing CIL methods have not considered the same class in different domains, resulting in a forgetting problem due to the classifier weight overwriting while struggling with intra-class domain confusion. Also, existing DIL methods suffer from inter-domain class confusion based on their consideration of increasing only the domains. To deal with this problem, we propose a simple but effective Incremental Classifier (IC) (Figure \ref{fig:ic}) which increments the final output node of the classifier layer if needed. Unlike the existing expandable methods, the proposed IC deals with the problem through a decision process on whether to increase the node using class-wise dynamic thresholding. Detailed descriptions are as follows.

As mentioned, the fully domain-specific classifier has problems, and it is necessary to set the appropriate criteria and increase nodes accordingly. To this end, we utilize the accuracy of each class in the learned domains and the current domain to determine the class-wise threshold ${\delta}_i$ dynamically as follows:
\begin{equation}
    \label{eq:thresholding}
    {\delta}_i = tanh(p_i), \:\:\:\:\: p_i=\gamma * \frac{\frac{1}{\left|D^{prev}\right|} {\Bigl(\sum_{d \in D^{prev}} Acc\left(C_i^d\right)\Bigl)}- Acc\left(C_i^{d_{new}}\right)}{\frac{1}{\left|D^{prev}\right|} \sum_{d \in D^{prev}} Acc\left(C_i^d\right)},
\end{equation}
where $D^{prev}=\{ d \:| 1\leq d \leq D \}$ (sequential integer set), $D$ refers to the number of previously learned tasks that contain the same classes as the newly arrived but different domains, $Acc(C_i^d)$ refers to the accuracy of $i$-th class in $d$-th domain, and $\gamma$ is a scaling factor. We consider only the classes that need to be learned in the current task in the entire processes of the IC (except for the shaded nodes in Figure \ref{fig:ic}). In this context, $p_i$ in Equation \ref{eq:thresholding} which is obtained by comparing the accuracies between the new domain and the already learned domains, represents the difficulty of the corresponding class in the new domain, and the classes whose accuracies are below the thresholds are considered as challenging classes to learn using the existing classifier. Therefore, the model increments its output nodes corresponding to each challenging class. During training, for challenging classes, only the logits of the increased nodes in the current task are used, and the existing nodes are not used to prevent forgetting (red solid line box in Figure \ref{fig:ic}). 

For classes that are not relatively difficult, \ie whose nodes do not need to be increased in the current task, the method for handling nodes is different from the above. In the case of classes that have not been increased and therefore have only a single corresponding node (the corresponding node to class $C_{k+1}^{1}$ in Figure \ref{fig:ic}), their single logits are used. In contrast, for the classes that have multiple nodes corresponding to one class which are not increased in the current task, but increased in previous tasks, a selection process is required to obtain a unique logit for learning.  Our model selects the maximum logit from multiple nodes (blue dotted line box with \textit{Max} operation in Figure \ref{fig:ic}).
We adopt the reasonable design choices for simple \textit{Max} operation as follows. A classifier learned with the cross-entropy (CE) loss can be regarded as an energy model \cite{lecun2006tutorial}, that aims to minimize the energy of it.
% The goal of the energy model is to minimize the energy of the model on the learning data, which equals to minimize  CE loss in a classifier. 
% Therefore, the CE loss also minimizes the energy of the classifier on the learning data. 
Existing energy-based methods \cite{liu2020energy,tang2021codes} have shown that data from in-distribution usually have lower energy than data from out-of-distribution for a certain classifier. 
In this context, the nodes with smaller energy for a single class are better suited to current training data than the others in that they are more likely to be in-distribution \cite{choe2024open,seo2024generative}. 
If the energy function is defined as $E(\boldsymbol{x}, y) = -f_y(\boldsymbol{x})$ (unnormalized negative logit of class $y$ for input $\boldsymbol{x}$) according to \cite{liu2020energy}, it can be determined that the node that produces the maximum logit (minimum energy) is the most suited for learning the current training data. 
Therefore, choosing the maximum logit for the final prediction is a simple but effective solution. This node selection strategy is also used in the inference. 

Finally, the CE loss ($\mathcal{L}_{CE}$) is applied to all final output logits (see Figure \ref{fig:ic}). Moreover, to prevent forgetting, we distill the knowledge of the nodes learned in the $(t-1)$-th task into nodes not selected by the \textit{Max} operation. Here, the corresponding logit and Kullback-Leibler divergence loss ($\mathcal{L}_{KL}$) are used (see Figure \ref{fig:ic}). The total loss of the proposed IC ($\mathcal{L}_{IC}$) is as follows:
\begin{equation}\label{eq:ic_loss}
\begin{aligned}
\mathcal{L}_{IC}= \mathcal{L}_{CE}(O^t, y)  +  \alpha \mathcal{L}_{KL}(O^t, O^{t-1}),
\end{aligned}
\end{equation}
where $O^t = f_\phi^{t, \mathcal{P}}(f_{\theta, A}(\boldsymbol{x})[CLS])$ refers to the final logits obtained by a classifier that is updated in current task $t$. $f_\phi^{t, \mathcal{P}}$ means the classifier of task $t$ with a function $\mathcal{P}$ that decides whether to increase the node with class-wise dynamic thresholding, and $\mathcal{P}$ is applied to the final output of the classifier.
% that the decision process on whether to increase the node using class-wise dynamic thresholding and appropriate usage mentioned above $\mathcal{P}$ is applied to the final output process of $f_\phi^{t, \mathcal{P}}$. 
$f_{\theta, A}$ and $f_{\phi}$ indicate the frozen ViT ($\theta$) with trainable adapters ($A$) and the classifier, respectively.  
$f_{\theta, A}(\boldsymbol{\cdot})[CLS]$ is the CLS token of the output after passing all transformer layers. $y$ is the label and $\alpha$ is the balancing weight between two objectives. 

\subsection{Training Objective}\label{sec:3.4}
\noindent Along with $\mathcal{L}_{CAST}$ in Equation \ref{eq:cast} and $\mathcal{L}_{IC}$ in Equation \ref{eq:ic_loss}, our end-to-end full optimization is as follows:
\begin{equation}\label{eq:total_loss}
\begin{aligned}
\mathcal{L}_{Total} = \beta\mathcal{L}_{CAST}(O^t, y) + \mathcal{L}_{IC}(O^t, O^{t-1}, y, \alpha),
\end{aligned}
\end{equation}
where $(\boldsymbol{x}, y) \in D_t$, data of current task $D_t$. While the parameters of the ViT are frozen, only the parameters $A$ of the adapters and $\phi$ of the classifier are updated. The final logits $O^t = f_\phi^{t, \mathcal{P}}(f_{\theta, A}(\boldsymbol{x})[CLS])$ same as in Equation \ref{eq:ic_loss}.
The proposed IC can solve the forgetting problem derived from the classifier weight overwriting and inter-domain class confusion that the existing methods suffered, and its superiority is demonstrated in the next (Section \ref{sec:experiment}).
\section{Experiments}
\label{sec:experiment}
\noindent In this section, we compared and evaluated our approach with state-of-the-art methods on widely used datasets. First, we introduce the experimental setup including the datasets, comparison baselines, and metrics in Sec. \ref{sec4.1}. The details about implementation and training are described in supplementary material. Also, we show extensive results of our experiments in Sec. \ref{sec4.2} to demonstrate the effectiveness of our approach.
Moreover, we conducted elaborate analysis including ablation studies in Sec. \ref{sec4.3} to interpret our approach in detail.

\begin{table*}[t!]
\centering
\caption{Dataset composition and configuration for each scenario. All IL scenarios were configured not deviate from the original composition of each dataset. Values with asterisk (*) refer to the number of domains on trainset pre-defined in the CORe50 \cite{lomonaco2017core50}. $N$ and $C_{t}$ indicates that number of tasks and classes per task respectively.}
\renewcommand{\arraystretch}{0.8}
\renewcommand{\tabcolsep}{1.5mm}
\resizebox{.8\textwidth}{!}{% 
\begin{tabular}{lccccccccc}
\specialrule{.1em}{.1em}{.1em} 
\multirow[b]{1.8}{*}{Dataset} & \multicolumn{2}{c}{Composition} & \multicolumn{2}{c}{CIL} & \multicolumn{2}{c}{DIL} & \multicolumn{2}{c}{VIL} \\
\cmidrule(lr){2-3} \cmidrule(lr){4-5} \cmidrule(lr){6-7} \cmidrule(lr){8-9} 
 &  \#Class & \#Domain & $N$ & $C_{t}$ & $N$ & $C_{t}$  & $N$ & $C_{t}$ \\
\specialrule{.05em}{.1em}{.1em} 
iDigits \cite{volpi2021continual} & 10 & 4 & 5 & 2 & 4 & 10             & 20 & 2 \\
CORe50 \cite{lomonaco2017core50} & 50 & 11 & 5 & 10 & 8* & 50 & 40 & 10 \\
DomainNet \cite{peng2019moment}& 345 & 6 & 5 & 69 & 6 & 345 & 30 & 10 \\
\specialrule{.1em}{.1em}{.1em} 
\end{tabular} %
}
\label{tab:config}
\end{table*}

\subsection{Experimental Setup}\label{sec4.1}
\noindent\textbf{Datasets.} We conducted experiments on three benchmarks, including iDigits \cite{volpi2021continual}, CORe50 \cite{lomonaco2017core50} and DomainNet \cite{peng2019moment} which are possible to construct IL scenarios that can cause a large shift in distribution by clearly distinguishing both classes and domains. The composition and configuration for each scenario about the datasets are described in Table \ref{tab:config} and please refer to the supplementary material for a more detailed explanation of the datasets.

\noindent\textbf{Comparison Baselines.} We compared ICON against naive baselines and various IL methods including the latest ones. 
% We carefully composed comparison methods and conducted all the experiments in the same environment such as backbone (ImageNet pre-trained ViT-B/16 \cite{dosovitskiy2020image}), single GPU (NVIDIA RTX 3090), implementation library (PyTorch  \cite{paszke2019pytorch}), \etc, for fair and exhaustive comparison. 
First, we set the \textit{Lower-bound} as usual supervised sequential fine-tuning result (notated as Fine-tuning in Table \ref{tab:main}). Then, we compared our proposed ICON with the regularization-based methods EWC \cite{kirkpatrick2017overcoming} and LwF \cite{li2016learning}. Moreover, we compared it with the recent prompt-based methods, including S-Prompts \cite{wang2022sprompts}, L2P \cite{wang2022learning}, DualPrompt \cite{wang2022dualprompt}, CODA-Prompt \cite{smith2023coda} and LAE \cite{gao2023unified}.

% In sequence, we compared ICON with regularization-based methods \cite{kirkpatrick2017overcoming,li2016learning} and prompt-based methods \cite{wang2022sprompts,wang2022learning,wang2022dualprompt,smith2023coda,gao2023unified} as follows.

% \noindent\textbf{Regularization-based Methods}. EWC \cite{kirkpatrick2017overcoming} and LwF \cite{li2016learning} are representative regularization-based methods. Since they are classic but extremely fundamental, which are mostly used as a comparison of incremental learning scenarios. 

% \noindent\textbf{Prompt-based Methods.} We also compare ICON with prompt-based methods. S-Prompts \cite{wang2022sprompts} targets DIL using prompts. Prompt-based methods also include L2P \cite{wang2022learning}, DualPrompt \cite{wang2022dualprompt}, CODA-Prompt \cite{smith2023coda} and LAE \cite{gao2023unified}, which have shown strong performance mainly in CIL.
% also leverage frozen pre-trained backbone and parameter efficient modules such as prompt.

\noindent\textbf{Evaluation Metrics.}  We evaluated the methods by the widely used two IL metrics: Average Accuracy which is the higher the better (marked as Avg. Acc↑), and Forgetting which is the lower the better. 
For the scenarios with clear task boundaries, we reported the final test score following the general protocol \cite{wang2022sprompts,wang2022learning,wang2022dualprompt,smith2023coda}.

% \begin{equation}
% \text{Average Accuracy:    }A_{T}=\frac{1}{T} \sum_{i=1}^{T} a_{T, i}
% \end{equation}
% where $T$ is the total number of tasks seen so far, and $a_{n, i}$ is the test accuracy on task $i$ after training the $n^{\text {th }}$ task. 

% \begin{equation}
% \text{Forgetting:    }F_{T}=\frac{1}{T-1} \sum_{i=1}^{T-1} f_{T, i}
% \end{equation}
% where $f_{j, i}$ is a measure of forgetting on task $i$ after training task $j$. $f_{j, i}$ is defined as the difference between best accuracy achieved on task $i$ in the past and the final accuracy of task $i$ after training task $j$ :
% \begin{equation}
% f_{j, i}=\max _{k \in\{1, \cdots, j-1\}} (a_{k, i}-a_{j, i})
% \end{equation}

% Regarding the threshold ${\delta}$ for decision of which classes should be incremented in the output nodes, we used ${\delta=0}$ for DomainNet, ${\delta=0.4}$ for CORe50, and ${\delta=0.5}$ for iDigits. Experimental results with regard to the ${\delta}$ can be seen in \ref{sec4.3}. The number of clusters after training each task is fixed to 2 for iDigits, and 3 for DomainNet and CORe50. Experimental results about the number of clusters are demonstrated in \ref{sec4.3} as well.

\begin{table*}[t!]
\caption{Main results with all of the IL scenarios. Experiments were conducted based on the latest incremental learning models. We used the bold and the underline as brief indications of the best and the second best, respectively.}
\resizebox{\textwidth}{!}{% 
\begin{NiceTabular}{cccccccccc@{}}
\specialrule{.1em}{.1em}{.1em} 
\multirow[b]{1.8}{*}{Method} & 
\multicolumn{2}{c}{CIL} & \multicolumn{2}{c}{DIL}  & \multicolumn{2}{c}{\cellcolor[HTML]{EEEEEE}VIL} & \multirow[b]{1.8}{*}{Average}\\ 
\cmidrule(lr){2-3} \cmidrule(lr){4-5} \cmidrule(lr){6-7} 
& Avg. Acc↑ & Forgetting↓ & Avg. Acc↑ & Forgetting↓ &\cellcolor[HTML]{EEEEEE} Avg. Acc↑ & \cellcolor[HTML]{EEEEEE} Forgetting↓ \\
\specialrule{.05em}{.1em}{.1em} 
&&&\multicolumn{2}{c}{\textbf{iDigits}}&&&& \\ 
\specialrule{.05em}{.1em}{.1em} 
Fine-tuning & 30.32$\pm$0.77 & 48.01$\pm$0.72 & 33.04$\pm$0.89 & 23.23$\pm$0.74 & \cellcolor[HTML]{EEEEEE}19.89$\pm$0.82 & \cellcolor[HTML]{EEEEEE}57.17$\pm$1.28 & 26.22$\pm$1.92 \\ 
EWC \cite{kirkpatrick2017overcoming} & 34.16$\pm$0.32 & 38.72$\pm$0.59 & 68.62$\pm$0.92 & 25.94$\pm$0.98  & \cellcolor[HTML]{EEEEEE}21.86$\pm$1.45 & \cellcolor[HTML]{EEEEEE}53.98$\pm$1.28 & 37.36$\pm$1.87 \\ 
LwF \cite{li2016learning}  & 39.88$\pm$0.91 & 33.35$\pm$0.52 & 69.61$\pm$0.33 & 25.81$\pm$0.69  & \cellcolor[HTML]{EEEEEE}23.44$\pm$0.14 & \cellcolor[HTML]{EEEEEE}53.65$\pm$0.42 & 41.91$\pm$2.19 \\ 
L2P \cite{wang2022learning} & 63.17$\pm$0.88 & 28.53$\pm$0.81 & 73.83$\pm$0.26 & 23.43$\pm$0.65 & \cellcolor[HTML]{EEEEEE}59.07$\pm$3.01 & \cellcolor[HTML]{EEEEEE}15.82$\pm$2.64 & 64.43$\pm$2.44 \\ 
S-Prompts \cite{wang2022sprompts} & 55.09$\pm$3.27 & 25.61$\pm$1.62 & 75.11$\pm$2.31 & 25.66$\pm$6.23  & \cellcolor[HTML]{EEEEEE}39.73$\pm$3.40 & \cellcolor[HTML]{EEEEEE}\underline{15.41$\pm$1.16} & 54.33$\pm$4.64 \\ 
DualPrompt \cite{wang2022dualprompt} & 68.82$\pm$0.97 & \textbf{11.81$\pm$1.77} & 76.42$\pm$0.46 & 26.33$\pm$0.62 & \cellcolor[HTML]{EEEEEE}60.25$\pm$2.92 & \cellcolor[HTML]{EEEEEE}23.40$\pm$3.50 & 67.61$\pm$3.43\\ 
CODA-P \cite{smith2023coda}  & \underline{69.97$\pm$1.02} & 19.83$\pm$2.28 & 77.42$\pm$0.71 & 22.20$\pm$0.18 & \cellcolor[HTML]{EEEEEE}\underline{63.30$\pm$3.08} & \cellcolor[HTML]{EEEEEE}16.43$\pm$2.63 & \underline{70.95$\pm$3.91} \\ 
LAE \cite{gao2023unified} & 65.77$\pm$0.83 & 28.47$\pm$0.77 & \underline{79.09$\pm$1.03} & 21.86$\pm$0.40 & \cellcolor[HTML]{EEEEEE}59.34$\pm$0.95 & \cellcolor[HTML]{EEEEEE}29.32$\pm$1.72 & 68.12$\pm$3.12 \\ 
\Hline[tikz={dashed}] 
\textbf{ICON (Ours)} & \textbf{71.53$\pm$0.68} & \underline{19.36$\pm$1.17} & \textbf{84.83$\pm$0.51} & \textbf{12.67$\pm$0.61} & \cellcolor[HTML]{EEEEEE} \textbf{75.11$\pm$2.39}& \cellcolor[HTML]{EEEEEE} \textbf{9.13$\pm$1.88} & \textbf{77.15$\pm$1.19}\\
\specialrule{.05em}{.1em}{.1em} 
&&&\multicolumn{2}{c}{\textbf{CORe50}}&&&& \\ 
\specialrule{.05em}{.1em}{.1em} 
Fine-tuning & 21.54$\pm$1.91 & 74.05$\pm$1.31 & 23.52$\pm$0.26 & 3.09$\pm$0.11 & \cellcolor[HTML]{EEEEEE}14.04$\pm$0.50 & \cellcolor[HTML]{EEEEEE}58.59$\pm$0.83 & 19.86$\pm$1.28 \\ 
EWC \cite{kirkpatrick2017overcoming} & 33.89$\pm$0.83 & 50.18$\pm$0.30 & 73.86$\pm$0.38 & 1.09$\pm$0.12 & \cellcolor[HTML]{EEEEEE}43.20$\pm$0.71 & \cellcolor[HTML]{EEEEEE}9.56$\pm$0.46 & 50.62$\pm$1.94 \\ 
LwF \cite{li2016learning}  & 34.53$\pm$0.55 & 41.05$\pm$0.30 & 74.35$\pm$0.52 & 0.81$\pm$0.27 &  \cellcolor[HTML]{EEEEEE}45.77$\pm$1.03 & \cellcolor[HTML]{EEEEEE}10.53$\pm$0.79 & 52.19$\pm$1.82 \\ 
L2P \cite{wang2022learning} & 70.03$\pm$0.51 & 6.51$\pm$0.59 & 80.72$\pm$0.39 & 0.51$\pm$0.28 &  \cellcolor[HTML]{EEEEEE}64.85$\pm$0.92 & \cellcolor[HTML]{EEEEEE}6.62$\pm$0.19 & 70.18$\pm$0.68 \\
S-Prompts \cite{wang2022sprompts} & 68.27$\pm$3.92 & 11.79$\pm$0.24 & \underline{86.50$\pm$0.46} & 0.92$\pm$0.31 &  \cellcolor[HTML]{EEEEEE}52.88$\pm$0.85 & \cellcolor[HTML]{EEEEEE} \underline{6.18$\pm$0.83} & 67.51$\pm$1.67 \\ 
DualPrompt \cite{wang2022dualprompt} & 71.96$\pm$0.37 & \underline{5.04$\pm$0.71} & 81.41$\pm$0.22 & $0.21\pm$0.76 &  \cellcolor[HTML]{EEEEEE}66.21$\pm$1.76 & \cellcolor[HTML]{EEEEEE} 7.20$\pm$0.88 & 71.46$\pm$1.00 \\
CODA-P \cite{smith2023coda}  & \underline{77.85$\pm$0.44} & \textbf{4.78$\pm$0.37} & 84.36$\pm$1.04 & 0.64$\pm$0.14 &  \cellcolor[HTML]{EEEEEE}69.28$\pm$0.24 & \cellcolor[HTML]{EEEEEE} 6.77$\pm$0.38 & 74.52$\pm$0.68 \\ 
LAE \cite{gao2023unified} & 77.11$\pm$0.31&18.38$\pm$1.67&83.09$\pm$0.71& \underline{0.17$\pm$0.51} & \cellcolor[HTML]{EEEEEE}\underline{77.11$\pm$1.37} & \cellcolor[HTML]{EEEEEE}8.23$\pm$2.59 & \underline{75.89$\pm$1.00}\\ 
\Hline[tikz={dashed}] 
\textbf{ICON (Ours)} & \textbf{80.85$\pm$0.23} & 7.68$\pm$0.52 & \textbf{89.01$\pm$0.33} & \textbf{0.17$\pm$0.21} & \cellcolor[HTML]{EEEEEE} \textbf{83.18$\pm$1.21} &\cellcolor[HTML]{EEEEEE}\textbf{4.72$\pm$0.24} & \textbf{84.34$\pm$0.59}\\
\specialrule{.05em}{.1em}{.1em} 
&&&\multicolumn{2}{c}{\textbf{DomainNet}}&&&& \\ 
\specialrule{.05em}{.1em}{.1em} 
Fine-tuning&35.43$\pm$0.58&47.79$\pm$0.28&39.52$\pm$0.32&28.81$\pm$0.64&\cellcolor[HTML]{EEEEEE}20.35$\pm$0.72&\cellcolor[HTML]{EEEEEE}43.22$\pm$1.14&31.66$\pm$0.57\\ 
EWC \cite{kirkpatrick2017overcoming}&53.04$\pm$0.53&24.41$\pm$0.48&41.58$\pm$0.26& 26.79$\pm$0.15&\cellcolor[HTML]{EEEEEE}36.68$\pm$0.25&\cellcolor[HTML]{EEEEEE}27.68$\pm$0.91&44.28$\pm$1.19\\ 
LwF \cite{li2016learning}&53.79$\pm$0.61&19.41$\pm$0.11&43.74$\pm$0.27&18.23$\pm$0.10&\cellcolor[HTML]{EEEEEE}38.17$\pm$0.35&\cellcolor[HTML]{EEEEEE}21.87$\pm$0.64&44.70$\pm$1.09\\ 
L2P \cite{wang2022learning}&60.90$\pm$0.69&\underline{8.23$\pm$0.90}&48.55$\pm$0.81&19.71$\pm$1.29& \cellcolor[HTML]{EEEEEE}48.98$\pm$0.69&\cellcolor[HTML]{EEEEEE}14.71$\pm$1.07&54.22$\pm$0.87\\
S-Prompts \cite{wang2022sprompts}&39.78$\pm$0.62&19.29$\pm$1.04&50.80$\pm$0.63&\textbf{4.20$\pm$0.53}& \cellcolor[HTML]{EEEEEE}35.90$\pm$0.54&\cellcolor[HTML]{EEEEEE} \underline{14.25$\pm$15.66}&42.54$\pm$1.21\\ 
DualPrompt \cite{wang2022dualprompt}&62.55$\pm$0.92&\textbf{7.62$\pm$1.07}&\underline{51.33$\pm$0.10}&\underline{9.60$\pm$1.41}&\cellcolor[HTML]{EEEEEE}49.36$\pm$1.05&\cellcolor[HTML]{EEEEEE}16.79$\pm$1.17 &56.00$\pm$0.84\\
CODA-P \cite{smith2023coda} &\underline{65.21$\pm$0.24}&15.01$\pm$0.21&49.13$\pm$0.83 
&25.96$\pm$1.13& \cellcolor[HTML]{EEEEEE}\underline{49.45$\pm$1.27}&\cellcolor[HTML]{EEEEEE}17.01$\pm$2.37&\underline{58.73$\pm$0.93}\\ 
LAE \cite{gao2023unified}&65.06$\pm$0.18&9.68$\pm$0.84&44.67$\pm$0.62&28.99$\pm$0.64& \cellcolor[HTML]{EEEEEE}49.01$\pm$1.18& \cellcolor[HTML]{EEEEEE}21.20$\pm$1.33&55.26$\pm$1.63\\ 
\Hline[tikz={dashed}] 
\textbf{ICON (Ours)}&\textbf{65.43$\pm$0.15}&9.72$\pm$0.46&\textbf{54.44$\pm$0.21}&13.32$\pm$0.46&\cellcolor[HTML]{EEEEEE} \textbf{53.37$\pm$0.47} &\cellcolor[HTML]{EEEEEE} \textbf{11.25$\pm$0.18}&\textbf{59.74$\pm$1.06}\\
\specialrule{.1em}{.1em}{.1em} 
\end{NiceTabular} %
}
\label{tab:main}
\end{table*}

\subsection{Experimental Results}\label{sec4.2}
\noindent\textbf{Main Results.} We conducted extensive experiments, including traditional IL scenarios as well as the proposed VIL scenario. The results are summarized in Table \ref{tab:main}. As shown in the table, ICON significantly outperformed the existing methods in our proposed VIL scenario in terms of both average accuracy and forgetting (shaded column). Moreover, we demonstrated the effectiveness of the proposed methods even in existing scenarios, resulting in the best average performance for all scenarios (unshaded columns). Furthermore, the existing state-of-the-art model showed a rather unstable performance (high standard deviation) in the two aforementioned scenarios, whereas ICON showed a very stable performance (low standard deviation). This indicates that while existing methods have limits in their learning abilities based on the order of class and domain that consists of each task, but ICON can reliably learn in any order. 

\noindent\textbf{Results on Cross-Domain Incremental Learning.}
Moreover, we conducted experiments on cross-domain incremental learning as shown in Table \ref{tab:cdil_table}. In cross-domain IL, both class and domain always increase at the same time in the following tasks, which is a subset of the VIL scenario. For all datasets, the number of classes in a task is the same as CIL setting. The number of tasks is 4 in iDigits, and 5 for CORe50 and DomainNet.
The outstanding performance on cross-domain IL implies that our proposed CAST and IC successfully accumulate knowledge from inputs on various sequences of data distributions. 

\begin{table*}[!t]
\caption{Results on Cross-Domain Incremental Learning scenario.}
\centering
\resizebox{\textwidth}{!}{% 
\begin{NiceTabular}{ccccccccc@{}}
\specialrule{.1em}{.1em}{.1em}
\multirow[b]{1.8}{*}{Method} & \multicolumn{2}{c}{\multirow{1}{*}{iDigits}} & \multicolumn{2}{c}{\multirow{1}{*}{CORe50}} & \multicolumn{2}{c}{\multirow{1}{*}{DomainNet}} & \multirow[b]{1.8}{*}{Average} \\ 
% \cmidrule(lr){1-2} 
\cmidrule(lr){2-3} \cmidrule(lr){4-5} \cmidrule(lr){6-7}
 &  \multicolumn{1}{c}{Avg. Acc↑} & \multicolumn{1}{c}{Forgetting↓} & \multicolumn{1}{c}{Avg. Acc↑}& \multicolumn{1}{c}{Forgetting↓} &\multicolumn{1}{c}{Avg. Acc↑}& \multicolumn{1}{c}{Forgetting↓} & \multirow{1}{*}{}\\

\specialrule{.05em}{.1em}{.1em} 
\multicolumn{1}{c}{ Fine-tuning} &   \multicolumn{1}{c}{21.62$\pm$5.21} & \multicolumn{1}{c}{51.01$\pm$6.86} &  \multicolumn{1}{c}{20.35$\pm$2.46} & \multicolumn{1}{c}{33.89$\pm$1.57} & \multicolumn{1}{c}{31.35$\pm$0.68}&\multicolumn{1}{c}{56.74$\pm$3.18}  & \multicolumn{1}{c}{24.44$\pm$2.78} \\
\multicolumn{1}{c}{EWC \cite{kirkpatrick2017overcoming}}&  \multicolumn{1}{c}{24.79$\pm$4.81} & \multicolumn{1}{c}{48.94$\pm$4.29} & \multicolumn{1}{c}{51.56$\pm$5.87} & \multicolumn{1}{c}{28.55$\pm$4.11} & \multicolumn{1}{c}{45.85$\pm$3.75} & \multicolumn{1}{c}{34.57$\pm$4.20} & \multicolumn{1}{c}{40.73$\pm$4.81} \\
\multicolumn{1}{c}{LwF \cite{li2016learning} }&   \multicolumn{1}{c}{34.71$\pm$7.38} & \multicolumn{1}{c}{36.34$\pm$4.91} & \multicolumn{1}{c}{54.12$\pm$5.18} & \multicolumn{1}{c}{27.10$\pm$1.41} & \multicolumn{1}{c}{43.12$\pm$3.14} & \multicolumn{1}{c}{33.13$\pm$2.77} & \multicolumn{1}{c}{43.98$\pm$5.23} \\
\multicolumn{1}{c}{L2P \cite{wang2022learning}}& \multicolumn{1}{c}{61.66$\pm$5.61} & \multicolumn{1}{c}{16.84$\pm$6.89} & \multicolumn{1}{c}{65.12$\pm$0.93} & \multicolumn{1}{c}{7.43$\pm$1.73} & \multicolumn{1}{c}{58.45$\pm$1.32} & \multicolumn{1}{c}{\underline{6.32$\pm$9.80}} & \multicolumn{1}{c}{61.74$\pm$2.62} \\
\multicolumn{1}{c}{S-Prompts \cite{wang2022sprompts}}& \multicolumn{1}{c}{47.40$\pm$9.61} & \multicolumn{1}{c}{\textbf{10.03$\pm$3.32}} & \multicolumn{1}{c}{62.41$\pm$1.47} & \multicolumn{1}{c}{11.87$\pm$4.27} & \multicolumn{1}{c}{38.82$\pm$1.76} & \multicolumn{1}{c}{9.12$\pm$2.11} & \multicolumn{1}{c}{49.54$\pm$4.28} \\
\multicolumn{1}{c}{DualPrompt \cite{wang2022dualprompt}}& \multicolumn{1}{c}{64.95$\pm$9.38} & \multicolumn{1}{c}{14.90$\pm$6.59} & \multicolumn{1}{c}{66.29$\pm$1.65} & \multicolumn{1}{c}{8.86$\pm$1.28} & \multicolumn{1}{c}{60.79$\pm$1.30} & \multicolumn{1}{c}{\textbf{5.34$\pm$1.94}} & \multicolumn{1}{c}{64.01$\pm$4.11} \\
\multicolumn{1}{c}{CODA-P \cite{smith2023coda}}& \multicolumn{1}{c}{\underline{73.09$\pm$10.85}} & \multicolumn{1}{c}{11.41$\pm$4.13} & \multicolumn{1}{c}{\underline{66.59$\pm$1.03}} & \multicolumn{1}{c}{\underline{6.08$\pm$0.95}} & \multicolumn{1}{c}{\underline{67.56$\pm$3.44}} & \multicolumn{1}{c}{10.47$\pm$1.71} & \multicolumn{1}{c}{\underline{69.08$\pm$5.10}} \\
\multicolumn{1}{c}{LAE \cite{gao2023unified}}& \multicolumn{1}{c}{68.24$\pm$9.68} & \multicolumn{1}{c}{19.22$\pm$4.22} & \multicolumn{1}{c}{66.28$\pm$1.64} & \multicolumn{1}{c}{10.17$\pm$1.79} & \multicolumn{1}{c}{61.78$\pm$4.56} & \multicolumn{1}{c}{17.16$\pm$3.05} & \multicolumn{1}{c}{65.43$\pm$5.29} \\
\Hline[tikz={dashed}] 
\multicolumn{1}{c}{\textbf{ICON (Ours)}}& \multicolumn{1}{c}{\textbf{75.73$\pm$5.63}} & \multicolumn{1}{c}{\underline{10.72$\pm$2.40}} & \multicolumn{1}{c}{\textbf{74.98$\pm$0.03}} & \multicolumn{1}{c}{\textbf{5.50$\pm$2.17}} & \multicolumn{1}{c}{\textbf{67.95$\pm$1.87}} & \multicolumn{1}{c}{8.18$\pm$1.80} & \multicolumn{1}{c}{\textbf{72.88$\pm$2.51}} \\

\specialrule{.1em}{.1em}{.1em} 
\end{NiceTabular} %
}
\label{tab:cdil_table}
\end{table*}

\begin{table*}[!t]
\caption{Ablation of CAST and IC in the VIL scenario.}
\centering
\resizebox{\textwidth}{!}{% 
\begin{NiceTabular}{cccccccccc@{}}
\specialrule{.1em}{.1em}{.1em}
\multicolumn{2}{c}{Method} & \multicolumn{2}{c}{\multirow{1}{*}{iDigits}} & \multicolumn{2}{c}{\multirow{1}{*}{CORe50}} & \multicolumn{2}{c}{\multirow{1}{*}{DomainNet}} & \multirow[b]{1.8}{*}{Average} \\ 
\cmidrule(lr){1-2} 
\cmidrule(lr){3-4} \cmidrule(lr){5-6} \cmidrule(lr){7-8}
\multicolumn{1}{c}{CAST} & \multicolumn{1}{c}{IC} & \multicolumn{1}{c}{Avg. Acc↑} & \multicolumn{1}{c}{Forgetting↓} & \multicolumn{1}{c}{Avg. Acc↑}& \multicolumn{1}{c}{Forgetting↓} &\multicolumn{1}{c}{Avg. Acc↑}& \multicolumn{1}{c}{Forgetting↓} & \multirow{1}{*}{}\\

\specialrule{.05em}{.1em}{.1em} 
\multicolumn{1}{c}{ } & \multicolumn{1}{c}{} &  \multicolumn{1}{c}{59.34$\pm$0.95} & \multicolumn{1}{c}{29.32$\pm$1.72} &  \multicolumn{1}{c}{77.11$\pm$1.37} & \multicolumn{1}{c}{8.23$\pm$2.59} & \multicolumn{1}{c}{49.01$\pm$1.18}&\multicolumn{1}{c}{21.20$\pm$1.33}  & \multicolumn{1}{c}{61.82$\pm$1.17} \\
\multicolumn{1}{c}{$\checkmark$} & \multicolumn{1}{c}{} & \multicolumn{1}{c}{68.34$\pm$2.09} & \multicolumn{1}{c}{18.30$\pm$6.41} & \multicolumn{1}{c}{79.20$\pm$0.59} & \multicolumn{1}{c}{\textbf{4.55$\pm$1.81}} & \multicolumn{1}{c}{50.56$\pm$0.51} & \multicolumn{1}{c}{17.50$\pm$0.96} & \multicolumn{1}{c}{66.03$\pm$1.06} \\
\multicolumn{1}{c}{} & \multicolumn{1}{c}{$\checkmark$}  & \multicolumn{1}{c}{66.97$\pm$1.03} & \multicolumn{1}{c}{14.32$\pm$5.10} & \multicolumn{1}{c}{81.13$\pm$3.01} & \multicolumn{1}{c}{5.32$\pm$2.11} & \multicolumn{1}{c}{51.60$\pm$1.32} & \multicolumn{1}{c}{11.90$\pm$0.77} & \multicolumn{1}{c}{66.57$\pm$1.79} \\
\Hline[tikz={dashed}] 
\multicolumn{1}{c}{$\checkmark$} & \multicolumn{1}{c}{$\checkmark$} & \multicolumn{1}{c}{\textbf{75.11$\pm$2.39}} & \multicolumn{1}{c}{\textbf{9.13$\pm$1.88}} & \multicolumn{1}{c}{\textbf{83.18$\pm$1.21}} & \multicolumn{1}{c}{4.72$\pm$0.24} & \multicolumn{1}{c}{\textbf{53.37$\pm$0.47}} & \multicolumn{1}{c}{\textbf{11.25$\pm$0.18}} & \multicolumn{1}{c}{\textbf{69.98$\pm$1.23}} \\

\specialrule{.1em}{.1em}{.1em} 
\end{NiceTabular} %
}
\label{tab:ablation}
\end{table*}

\subsection{Analysis}\label{sec4.3}
\noindent\textbf{Ablation Study.} We further investigated the effectiveness of each component of ICON in Table \ref{tab:ablation}. We conducted an ablation study starting from our baseline, adding CAST, IC, and both. We confirmed that using the CAST loss to regularize the direction of current learning by considering the shifts in parameters in history is effective in learning versatile tasks. Furthermore, we also confirmed that with IC, the model can accumulate knowledge of different domains within the same classes by increasing its output node when the existing nodes are decided not to be appropriate to be used in a new domain. By adopting CAST and IC for VIL, the model can leverage the effectiveness of each component with synergy. The ablation results showed that each component of ICON was effective in alleviating catastrophic forgetting in a situation where input tasks are erratic. 

Moreover, we conducted an ablation study on IC by dividing it into node expansion and distillation as shown in Table \ref{tab:ic_ablation}. Implementation of simple expansion of output nodes without distillation to preserve the knowledge showed considerable performance gain. It indicates that the separation of the output nodes corresponding to a single class can accommodate disparate knowledge from different domains. As already known, a distillation of knowledge from previous classifier was also helpful in mitigating catastrophic forgetting in VIL as well. 

\noindent\textbf{Number of Clusters.} We conducted our experiments with different numbers of clusters in our proposed CAST, varying from 0 to 6 in the VIL setting. We demonstrated the performance gain from our baseline for each dataset in Figure \ref{fig:k}, where not using CAST is equivalent to when the number of clusters is 0, and we used all previous shifts in the history for CAST loss when the number of clusters is 1.
As you can see in Figure \ref{fig:k}, iDigits gained the best performance when the number of clusters is 2 and CORe50 when 3. Since the number of the entire tasks is 20 for iDigits and 40 for CORe50, it can be interpreted that as the number of tasks grows, clustering the history shifts into a bigger number is effective when the sequence becomes longer. Using bigger than 3 for the number of clusters did not show any noticeable performance gain. 
% \begin{table}[!t]
\begin{minipage}{.499\textwidth}
\captionof{table}{Ablation of IC in iDigits. NE and KD refer to node expansion and knowledge distillation respectively.}
\centering
\renewcommand{\arraystretch}{1.1}
\renewcommand{\tabcolsep}{3.0mm}
\resizebox{\textwidth}{!}{
\begin{NiceTabular}{ccccc@{}}
\specialrule{.1em}{.1em}{.1em}
\multicolumn{1}{c}{NE} & \multicolumn{1}{c}{ KD }& \multicolumn{1}{c}{ Avg. Acc↑} & \multicolumn{1}{c}{ Forgetting↓} \\
\specialrule{.05em}{.1em}{.1em}
\multicolumn{1}{c}{} & \multicolumn{1}{c}{} &  \multicolumn{1}{c}{  59.34$\pm$0.95} & \multicolumn{1}{c}{ 29.32$\pm$1.72}  \\
\multicolumn{1}{c}{$\checkmark$} & \multicolumn{1}{c}{} & \multicolumn{1}{c}{  63.10$\pm$3.58} & \multicolumn{1}{c}{25.50$\pm$2.98}  \\
\multicolumn{1}{c}{} & \multicolumn{1}{c}{$\checkmark$}  & \multicolumn{1}{c}{ 64.66$\pm$3.10} & \multicolumn{1}{c}{ 19.30$\pm$3.27}   \\
\Hline[tikz={dashed}] 
\multicolumn{1}{c}{$\checkmark$} & \multicolumn{1}{c}{$\checkmark$} & \multicolumn{1}{c}{  \textbf{66.97$\pm$1.03}} & \multicolumn{1}{c}{  \textbf{14.32$\pm$5.10}}  \\
\specialrule{.1em}{.1em}{.1em} 
\end{NiceTabular} %
}
\label{tab:ic_ablation}
\end{minipage}
% \end{table}
\begin{minipage}{.499\textwidth}
% \begin{figure}[t!]
\centering
% \begin{subfigure}[b]{0.8\columnwidth}
      \includegraphics[width=.9\textwidth]{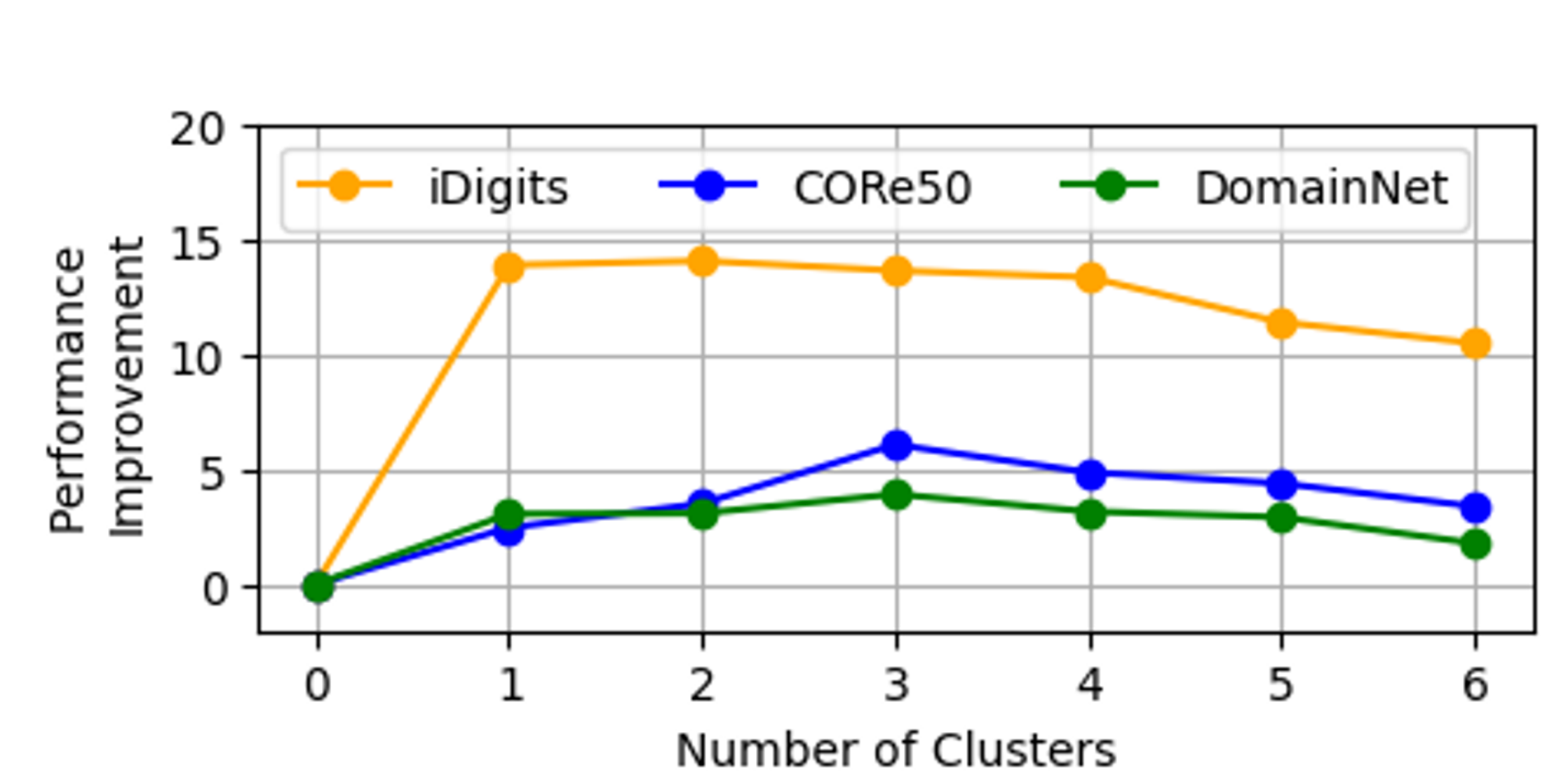}
      \captionof{figure}{Performance gain from number of clusters. }
      \label{fig:k}
      
% \end{subfigure}

% Not using CAST is equivalent to when the number of clusters is 0. We used all previous shifts in the history for CAST loss when the number of clusters is 1. }
\label{fig:k}
% \end{figure}
\end{minipage}

\section{Conclusion}
\label{sec:conclusion}
\noindent In this work, we proposed a new IL scenario named Versatile Incremental Learning (VIL), that reflects a more complex real-world derived from random incremental streams (classes, domains, or both) without any incremental prior knowledge. We defined the key challenges in VIL and proposed a novel framework, coined ICON (Incremental Classifier with Adaptation Shift cONtrol), composed of Cluster-based Adaptation Shift conTrol (CAST) loss and Incremental Classifier (IC). We demonstrated that ICON showed SOTA performance in the proposed VIL, as well as existing IL scenarios, and its effectiveness through various experiments. We look forward to our proposed VIL scenario serves as a new starting point for real-world IL field.

Nevertheless, there is still room for improvement in our work, especially regarding the scenario. A scenario that considers a varying number of classes and domains in a task can deal with a more realistic scenario, since in real-world, the distributions of classes and domains in a task can change. 

\clearpage
\section*{Acknowledgements}
This work was supported by MSIT (Ministry of Science and ICT), Korea, under the ITRC (Information Technology Research Center) support program (IITP-2024-RS-2023-00258649) supervised by the IITP (Institute for Information \& Communications Technology Planning \& Evaluation), and in part by the IITP grant through MSIT (Development of Semi-Supervised Learning Language Intelligence Technology and Korean Tutoring Service for Foreigners) under Grant 2019-0-00004, and in part by the IITP grant through MSIT (Artificial Intelligence Innovation Hub) under Grant 2021-0-02068, and by the IITP grant funded by the Korea government (MSIT) (No.RS-2022-00155911, Artificial Intelligence Convergence Innovation Human Resources Development (Kyung Hee University)).

{
    % \small
    \bibliographystyle{splncs04}
    % \bibliography{main}

}

\par\vfill\par
% Now we have reached the maximum length of an ECCV \ECCVyear{} submission (excluding references).
% References should start immediately after the main text, but can continue past p.\ 14 if needed.
\clearpage  % TODO REVIEW/FINAL: This \clearpage needs to be removed from both review and camera-ready versions.

% ---- Bibliography ----
%
% BibTeX users should specify bibliography style 'splncs04'.
% References will then be sorted and formatted in the correct style.
%

\end{document}

% ---------------------------------------------------------------
% TODO REVIEW: Replace with your title
\title{Versatile Incremental Learning: Towards Class and Domain-Agnostic Incremental Learning} 

% TODO REVIEW: If the paper title is too long for the running head, you can set
% an abbreviated paper title here. If not, comment out.
\titlerunning{Versatile Incremental Learning}

% TODO FINAL: Replace with your author list. 
% Include the authors' OCRID for the camera-ready version, if at all possible.
\author{
Min-Yeong Park\thanks{Equally contributed}\orcidlink{0009-0007-4143-7283} \and
Jae-Ho Lee\printfnsymbol{1}\orcidlink{0009-0003-0960-824X} \and
Gyeong-Moon Park\thanks{Corresponding author}\orcidlink{0000-0003-4011-9981}
}

% TODO FINAL: Replace with an abbreviated list of authors.
\authorrunning{M.-Y. Park et al.}
% First names are abbreviated in the running head.
% If there are more than two authors, 'et al.' is used.

% TODO FINAL: Replace with your institution list.
\institute{Kyung Hee University, Yongin, Republic of Korea \\
\email{\{pmy0792, jh.lee, gmpark\}@khu.ac.kr}}

\maketitle

\noindent In this supplementary material, we further validate the effectiveness of the proposed ICON by providing as follows:
\begin{itemize}
    \item Architecture Details. \ref{sec:architecture}
    \item Experimental Details. \ref{sec:experimental}
    \item Additional Experiments and Analysis. \ref{sec:additional}
    \item Additional Ablation Results. \ref{sec:ablation}
    \item Limitations and Future Work. \ref{sec:limitations}
    \item Hyperparameters of Comparison Models. \ref{sec:hyperparam_comparison}
\end{itemize}

\section{Architecture Details}\label{sec:architecture} 
\begin{figure}[!t]
\centering
\captionsetup[subfigure]{labelfont=footnotesize,textfont=footnotesize}
\begin{subfigure}[b]{0.539\columnwidth}
      \includegraphics[width=\textwidth]{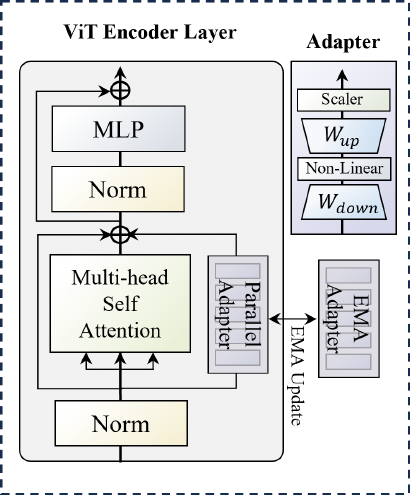}
      \caption{ViT layer with adapter.}
      \label{fig:architecture_deatail_1}
\end{subfigure}
\begin{subfigure}[b]{0.408\columnwidth}
      \includegraphics[width=\textwidth]{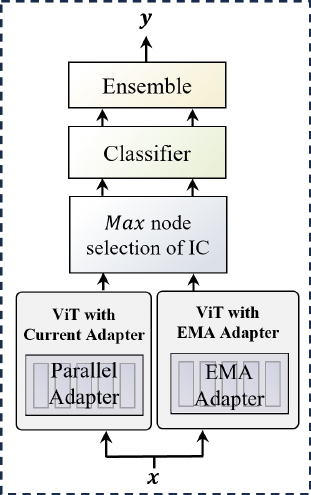}
      \caption{Ensemble in inference.}
      \label{fig:architecture_deatail_2}
\end{subfigure}
\caption{Detailed architecture visualization of proposed ICON.}
\label{fig:architecture}
\end{figure}

As shown in Figure \ref{fig:architecture}, our architecture has a very simple structure. Based on the typical ViT encoder layer \cite{dosovitskiy2020image}, each adapter is added parallel to the attention of the layers included in the adapter locations demonstrated in Table \ref{tab:config}. We also introduce trainable scaling parameters that have been proven its effectiveness in \cite{he2022towards}. As shown in Figure \ref{fig:architecture_deatail_1}, we adopt an Exponential Moving Average (EMA) to each adapter to preserve global knowledge, with our proposed CAST. We also apply an adapter ensemble mechanism (see Figure \ref{fig:architecture_deatail_2})) that takes the bigger logits from the model with the current adapter or from the model with the EMA adapter, to utilize global knowledge in the inference phase, following our baseline \cite{gao2023unified}.

\section{Experimental Details}\label{sec:experimental}
\subsubsection{Datasets.} We conducted experiments on three benchmarks, including iDigits \cite{volpi2021continual}, CORe50 \cite{lomonaco2017core50} and DomainNet \cite{peng2019moment} which are possible to construct an incremental learning scenario that can cause a large shift in distribution (both class and domain change) by clearly distinguishing both classes and domains. We follow \cite{volpi2021continual} to compose a digit recognition incremental scenario, which is composed of four datasets: MNIST \cite{lecun1998gradient}, SVHN \cite{netzer2011reading}, MNIST-M \cite{ganin2015unsupervised} and SynDigit \cite{ganin2015unsupervised}. Each dataset is treated as a different domain. CORe50 \cite{lomonaco2017core50} is a widely used dataset for domain incremental learning or continual real-world object recognition. It has 50 classes collected from a large variety of views in time, and each class has 11 distinct domains. In an incremental learning setting, the data from 8 domains are used for training and the data from the rest (\ie, unseen) 3 domains as a test set. DomainNet \cite{peng2019moment} is a very popular dataset for domain incremental learning or domain adaptation. All the data from the 6 distinct domains and it has large 345 classes for classification. Unlike the CORe50 \cite{lomonaco2017core50}, it includes not only real-world data but also data from unreal domains such as painting, clipart and infographic.

\subsubsection{Data Pre-processing.} We adopted a very simple data augmentation strategy for all experiments, following L2P \cite{wang2022learning}, DualPrompt \cite{wang2022dualprompt}, LAE \cite{gao2023unified}. First, input images are randomly resized to 224 $\times$ 224 with bilinear interpolation. Unlike the JAX \cite{jax2018github} implementation, scale$=$(0.08, 1.0) and ratio$=$(3/4, 4/3) were applied with random resized crop as default. Second, cropped images were randomly flipped horizontally. In the inference phase, the images are resized to 256 $\times$ 256, and subsequently center cropped to 224 $\times$ 224. We used the normalization to the range of [0, 1] as the last step of each augmentation strategy. 
% The PyTorch style code is provide in Algorithm \ref{alg:augmentation}. 

\subsubsection{Implementation Details.} We conducted all the experiments on a single NVIDIA GeForce RTX 3090 GPU. To make fair comparisons, we used standard ImageNet \cite{deng2009imagenet} pre-trained ViT-B/16\footnote{storage.googleapis.com/vit\_models/imagenet21k/ViT-B\_16.npz} \cite{dosovitskiy2020image} as a backbone of all methods. Furthermore, for the unification of the implementation library (PyTorch \cite{paszke2019pytorch}), we used the PyTorch implementation for L2P\footnote{github.com/JH-LEE-KR/l2p-pytorch} \cite{wang2022learning} and DualPrompt\footnote{github.com/JH-LEE-KR/dualprompt-pytorch} \cite{wang2022dualprompt}, which the official code is written in JAX \cite{jax2018github}. S-Prompts \cite{wang2022sprompts} suggests image S-Prompts (S-iPrompts) and language-image S-Prompts (S-liPrompts). We used S-iPrompts as a comparison model without using text features (using ViT as a backbone, not CLIP \cite{radford2021learning}), for fair comparison. 

\subsubsection{Training Details.} We trained the model for 5 epochs per task, 3 epochs for training only the classifier and the other 2 epochs for training both classifier and adapters while freezing other parts of pretrained ViT, following our baseline \cite{gao2023unified}. With the loss function for training, we adopted $\alpha$ to be 1 and $\beta$ to be different among datasets. We used $\beta =0.05$ for CORe50 and iDigits and ${\beta=0.01}$ for DomainNet. For thresholds in IC, we used $\gamma$ to be 2.

\subsubsection{Hyperparameters.} We summarize the hyperparameters for each dataset used in the main experiments (Table 3. of main paper) in Table \ref{tab:config}. In Table \ref{tab:config}, Warmup epochs means the number of epochs where the model except for the classifier is frozen, and the classifier and adapters are simultaneously trained after Warmup epochs are finished. The number of Clusters indicates the $K$ value of the K-Means algorithm used in the proposed CAST for clustering history shifts. The coefficient of distillation and CAST in the loss function are denoted as $\alpha$ and $\beta$. Adapter Location represents the index of the 12 ViT layers to which the adapter is added, and index 0 points to the first layer. We add five adapters sequentially and the hidden dimension size of the feature that is entered into each adapter to be downsampled is written as Adapter Downsample. Furthermore, in our proposed VIL scenario, extreme cases of training all classes sequentially and then re-training classes already learned in other domains (\ie., [CIL, CIL, CIL, ..., DIL, DIL, DIL]) can also occur, which can make the performance difference large. Therefore, we conducted experiments with various sequences (random seeds).

Our ICON does not use a prompt, but most comparison models use it. Since this is accompanied by relatively more hyperparameters than ours, such as prompt pool size, top-$K$, prompt length, \etc., we also provide the hyperparameters of the comparisons that were used in the experiments in Section \ref{sec:hyperparam_comparison}.

\begin{table*}[!t]
\caption{Detailed hyperparameters and experimental configuration of proposed ICON for each dataset.}
\renewcommand{\tabcolsep}{2mm}
\resizebox{\textwidth}{!}{%
\begin{tabular}{lccc}
\specialrule{.1em}{.1em}{.1em} 
\multicolumn{1}{l}{Configuration} & iDigits \cite{volpi2021continual} & CORe50 \cite{lomonaco2017core50} & DomainNet \cite{peng2019moment} \\ 
\specialrule{.05em}{.1em}{.1em} 
Optimizer & Adam \cite{Kingma2015Adam}& Adam \cite{Kingma2015Adam}& Adam \cite{Kingma2015Adam} \\
Base LR & 0.0028125 & 0.0028125 & 0.0028125 \\
Optimizer Betas & $\beta_{1}$, $\beta_{2}$ = 0.9, 0.999 & $\beta_{1}$, $\beta_{2}$ = 0.9, 0.999 & $\beta_{1}$, $\beta_{2}$ = 0.9, 0.999 \\
Batch Size & 24 & 24 & 24 \\
EMA Decay & 0.9999 & 0.9999 & 0.9999 \\
Total Epochs & 5 & 5 & 5 \\
Warmup Epochs & 3 & 3 & 3 \\
Number of Clusters & 2 & 3 & 3 \\
% IC Threshold & 0.5 & 0.4 & 0.0 \\
\alpha & 1 & 1 & 1 \\
\beta & 0.05 & 0.05 & 0.01 \\
Adapter Location & {[}0, 1, 2, 3, 4{]} & {[}0, 1, 2, 3, 4{]} & {[}0, 1, 2, 3, 4{]} \\
Adapter Downsample & 5 & 5 & 5 \\
% Random Seeds & {[}5722, 26521, 24728, 29031, 20527{]} & {[}1024, 1302, 7081, 19438, 29665{]} & {[}29690, 25862, 19418, 32443, 13185{]} \\ 
\specialrule{.1em}{.1em}{.1em} 
\end{tabular}%
}
\label{tab:config}
\end{table*}

\subsubsection{Evaluation Metrics.} We provide a formal definition of the evaluation metrics that were used in all experiments. Each metric is formally defined as follows:

\begin{equation}
\text{Average Accuracy:    }A_{T}=\frac{1}{T} \sum_{i=1}^{T} a_{T, i},
\end{equation}
where $T$ is the total number of tasks seen until the current task, and $a_{n, i}$ is the test accuracy on task $i$ after training the $n^{\text {th }}$ task. 
\begin{equation}
\text{Forgetting:    }F_{T}=\frac{1}{T-1} \sum_{i=1}^{T-1} f_{T, i},
\end{equation}
where $f_{j, i}$ is a measure of forgetting on task $i$ after training task $j$. $f_{j, i}$ is defined as the difference between the best accuracy achieved on task $i$ in the past and the final accuracy of task $i$ evaluated after training task $j$ :
\begin{equation}
f_{j, i}=\max _{k \in\{1, \cdots, j-1\}} (a_{k, i}-a_{j, i}),
\end{equation}
To validate the performance of the proposed ICON on all of the scenarios, we also took an average of Avg. Acc of all scenarios. It can be formulated as follows:
\begin{equation}
\text{Average:    }A_{Avg}=\frac{1}{|S|}\sum_{s=1}^{S} A_s,
\end{equation}
where $|S|$ is the total number of scenarios, and $A_s$ is the final average accuracy of scenario $s$. In this paper, we used $|S|$=3, for CIL, DIL, and VIL.

\section{Additional Experiments and Analysis}\label{sec:additional}
\begin{table}[!t]
\caption{Average accuracy on various number of tasks.}
\renewcommand{\arraystretch}{1.2}
\renewcommand{\tabcolsep}{0.5mm}
\resizebox{\columnwidth}{!}{% 
\begin{tabular}{lcccccccccc}
\specialrule{.1em}{.1em}{.1em} 
\multirow[b]{1.8}{*}{Method}  & \multicolumn{2}{c}{iDigits \cite{volpi2021continual}}  & \multicolumn{3}{c}{CORe50 \cite{lomonaco2017core50}} & \multicolumn{3}{c}{DomainNet \cite{peng2019moment}}\\
\cmidrule(lr){2-3} \cmidrule(lr){4-6} \cmidrule(lr){7-9}
& \multicolumn{1}{c}{8 Tasks} & \multicolumn{1}{c}{20 Tasks} & \multicolumn{1}{c}{16 Tasks} & \multicolumn{1}{c}{40 Tasks}&\multicolumn{1}{c}{80 Tasks} &\multicolumn{1}{c}{18 Tasks} & \multicolumn{1}{c}{30 Tasks} & \multicolumn{1}{c}{138 Tasks}   \\
\specialrule{.05em}{.1em}{.1em} 
LAE \cite{gao2023unified} &79.67&59.34&86.97&77.11&72.05&47.16&49.01&49.22 \\
\textbf{ICON (Ours)} & \textbf{81.59} & \textbf{75.11}&\textbf{88.04}& \textbf{83.18}&\textbf{76.34}&\textbf{48.95}&\textbf{53.37}&\textbf{50.84}\\
\specialrule{.1em}{.1em}{.1em} 
\end{tabular} %s
}
\label{tab:num_task}
\end{table}

\subsubsection{Robustness to Various Number of Tasks.} 
\begin{figure}[!t]
\centering
\begin{subfigure}[b]{0.9\columnwidth}
      \includegraphics[width=\textwidth]{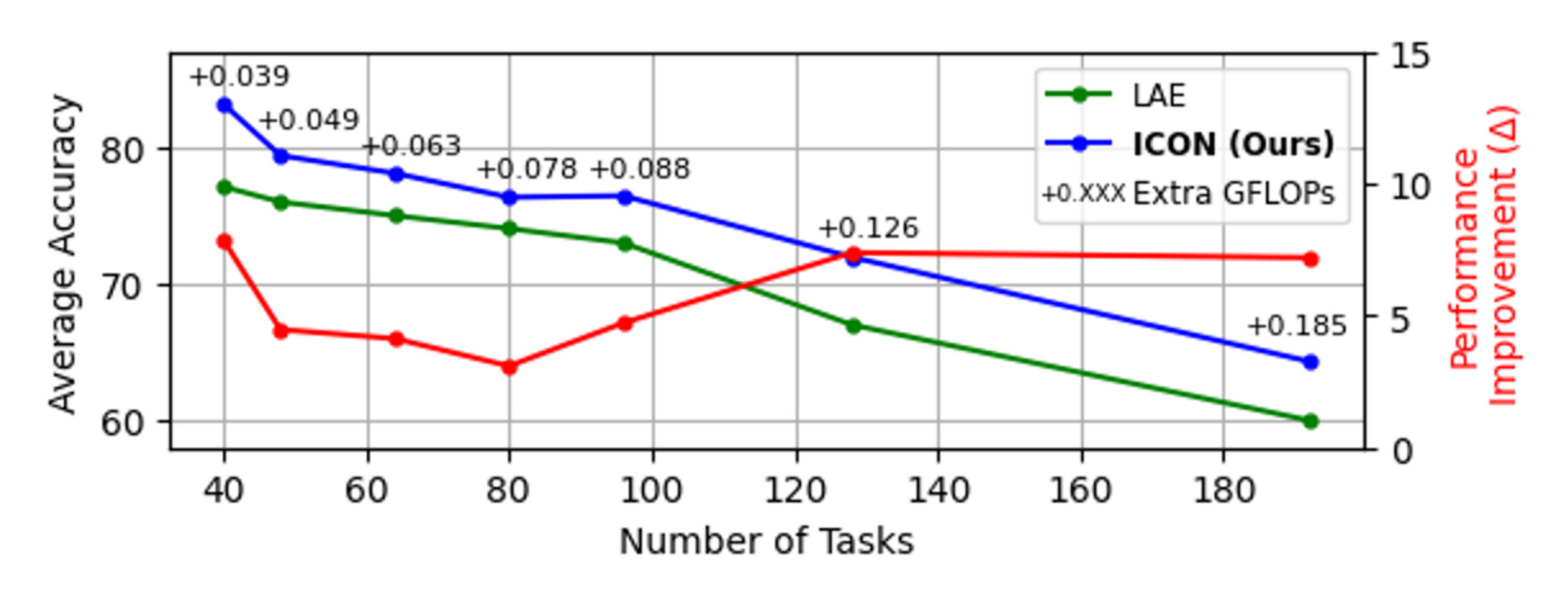}
\end{subfigure}
\captionsetup{font=footnotesize}
\caption{Average accuracy on a higher number of tasks in CORe50. Extra GFLOPs is obtained in comparison with our baseline LAE.}
\label{fig:num_task}
\end{figure}

We conducted additional experiments to validate the robustness of our method on various numbers of task sequences as shown in Table \ref{tab:num_task} with the results of standard configuration included in the main paper. The number of the entire task sequence for the VIL setting is dependent on the number of domains in it and the number of classes in a single task. For DomainNet, we conducted experiments in different task configurations so that the number of entire tasks becomes 18 and 138. Since the entire class existing in DomainNet is 345, the sequence of tasks become 18 by comprising a single task with 115 classes, and 138 by comprising a single task with only 23 classes. In the same way, we configured the number of tasks for CORe50 to be 16 and 80. 16 tasks for CORe50 are in the case of a single task with 25 classes, and 80 tasks are in the case of a single task with 5 classes. For iDigits, by having 5 classes in a single task, we can configure it to have 8 tasks entirely. As it can be seen in Table \ref{tab:num_task}, in iDigits and CORe50, as the number of tasks increases, the average accuracy degrades since catastrophic forgetting becomes severe. Also, ICON achieved noticeable performance improvements compared to our baseline LAE \cite{gao2023unified} for all results on various task numbers, showing that the effectiveness of ICON is robust for various number of tasks.
To thoroughly show the results on a more diverse number of tasks in CORe50, we visualize average accuracies in Figure \ref{fig:num_task}. As shown in the figure, the performance improvement of ICON compared to the baseline LAE (see red line) is consistent with the number of tasks, verifying that our proposed ICON is robust with regard to the number of tasks.
% \noindent\textbf{Analysis on Routing Nodes for a Single Class.} For IC to be effectively used for learning multiple domains corresponding to a single class, the network should decide which node to select among multiple logits for each class. To be specific, a node whose time point of initialization equals to task identity of inputs should be selected at test time during max pooling. For this sake, we examined the average accuracy of routing after learning each task as well as the number of increased nodes for each task in DomainNet. The overall average accuracy for the entire tasks are around slightly above 70\% for DomainNet, indicating that the network can select correct nodes properly. The trend of average accuracy of routing and the number of increased nodes per each task is shown in Figure \ref{fig:nodes}.
\begin{table}[!t]
\centering
\renewcommand{\arraystretch}{1.2}
\caption{Comparisons of joint training, the existing SOTA (LAE), and the proposed ICON in the VIL scenario.}
\renewcommand{\tabcolsep}{4mm}
\resizebox{0.9\textwidth}{!}{%
\begin{tabular}{lccc}
\specialrule{.1em}{.1em}{.1em} 
\multirow{2}{*}{Method} & \multicolumn{3}{c}{Dataset} \\ \cline{2-4} 
& \multicolumn{1}{c}{iDigits \cite{volpi2021continual}} & \multicolumn{1}{c}{CORe50 \cite{lomonaco2017core50}} & \multicolumn{1}{c}{DomainNet \cite{peng2019moment}}  \\ 
\specialrule{.05em}{.1em}{.1em} 
LAE \cite{gao2023unified}& \multicolumn{1}{c}{59.34$\pm$0.95} & \multicolumn{1}{c}{77.11$\pm$1.37} & 49.01$\pm$1.18 \\
\textbf{ICON (Ours)} & \multicolumn{1}{c}{75.11$\pm$2.39} & \multicolumn{1}{c}{83.18$\pm$1.21} & 53.37$\pm$0.47 \\ 
\specialrule{.05em}{.1em}{.1em} 
Joint Training  & \multicolumn{1}{c}{\textbf{87.18$\pm$0.13}} & \multicolumn{1}{c}{\textbf{91.66$\pm$0.25}} & \textbf{76.91$\pm$0.61}\\
\specialrule{.1em}{.1em}{.1em} 
\end{tabular}%
}
\label{tab:joint}
\end{table}
\begin{table}[!t]
\centering
\renewcommand{\tabcolsep}{2mm}
\caption{Comparison of computational cost on ViT-B/16.}
\resizebox{\textwidth}{!}{%
\begin{tabular}{lcccccc}
\specialrule{.1em}{.1em}{.1em} 
Method &L2P \cite{wang2022learning}&DualPrompt \cite{wang2022dualprompt}&CODA-P \cite{smith2023coda}&S-Prompts \cite{wang2022sprompts}& LAE \cite{gao2023unified}&\textbf{ICON (Ours)} \\ 
\specialrule{.05em}{.1em}{.1em} 
GFLOPs &116.27& 105.89& 140.12&75.32&\textbf{71.34}&\underline{71.52} \\ 
\specialrule{.1em}{.1em}{.1em} 
\end{tabular}%
}
\label{tab:cost}
\end{table}

\subsubsection{Result of Joint Training.} To demonstrate the gap between the proposed method and joint training results, we measured the joint training results as shown in Table \ref{tab:joint}. While the performances of our proposed ICON still have gaps compared to those of joint training, it is noteworthy that our ICON significantly narrowed the gaps by effectively dealing with the challenges in VIL, especially in iDigits.

\begin{table}[!t]
\centering
\captionof{table}{Experiments on different backbones in CORe50.}
\begin{tabular}{lcccc}
\specialrule{.1em}{.1em}{.1em} 
\multicolumn{1}{c}{\multirow{2}{*}{Dataset}} & \multicolumn{2}{c}{ViT-S/16 \cite{dosovitskiy2020image}} & \multicolumn{2}{c}{ViT-L/16 \cite{dosovitskiy2020image}} \\ \cline{2-3} \cline{4-5} 
\multicolumn{1}{c}{} & LAE \cite{gao2023unified} & \textbf{ICON (Ours)} & LAE \cite{gao2023unified}& \textbf{ICON (Ours)} \\ 
\specialrule{.05em}{.1em}{.1em} 
iDigits \cite{volpi2021continual}&53.40&\textbf{55.34}&69.68&\textbf{72.42}  \\
DomainNet \cite{peng2019moment}&41.80&\textbf{45.10}&51.80&\textbf{54.09} \\ 
\specialrule{.05em}{.1em}{.1em} 
GFLOPs &37.09&37.27&247.81&247.99 \\
\specialrule{.1em}{.1em}{.1em} 
\end{tabular}%
\label{tab:backbone}
\end{table}

\subsubsection{Computational Complexity and Scalability.} 
We analyze computational complexity in Table \ref{tab:cost}, which demonstrates that our method demands fewer or equivalent resources compared to the baselines, while achieving the best performances.
In addition, to investigate the scalability, we conducted experiments on various sizes of backbones and extreme numbers of tasks. As shown in Table \ref{tab:backbone}, the proposed ICON performed the best in all sizes of backbones with equivalent costs compared to the baseline LAE.
Moreover, even when extended to extreme numbers of tasks, the proposed ICON only required negligible extra costs as indicated in Figure \ref{fig:num_task}.
Therefore, our method is applicable in dynamic and large-scale scenarios without issues in terms of computational complexity and scalability.

\begin{minipage}{.45\textwidth}
    \centering
       % \begin{subfigure}{\columnwidth}
          \includegraphics[width=\textwidth]{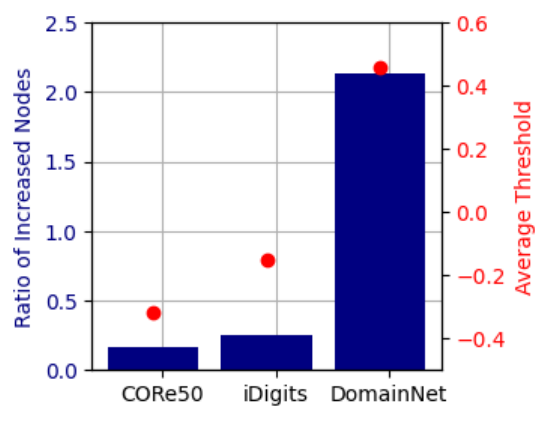}
          \label{fig:thre}
          \captionof{figure}{Average threshold and the number of increased nodes.}
        % \end{subfigure}}
        
\end{minipage}%
\begin{minipage}{.5\textwidth}
    % \begin{table}[t]
    \renewcommand{\tabcolsep}{3mm}
    \captionof{table}{\small Ablation of Dynamic Threshold (DT).}
    \centering
    \resizebox{0.9\columnwidth}{!}{% 
    \begin{tabular}{lccc}
    \specialrule{.1em}{.1em}{.1em} 
    \multirow{1}{*}{DT}  & \multicolumn{1}{c}{  iDigits \cite{volpi2021continual}  }  & \multicolumn{1}{c}{  CORe50 \cite{lomonaco2017core50}  } & \multicolumn{1}{c}{DomainNet \cite{peng2019moment}}\\
    \specialrule{.05em}{.1em}{.1em} 
        &74.79&82.71&52.99\\
      $\checkmark$  &\textbf{75.11}&\textbf{83.18}&\textbf{53.37} \\
    \specialrule{.1em}{.1em}{.1em} 
    \end{tabular} %
    }
    \label{tab:dt}
    % \end{table}
\end{minipage}%

\subsubsection{Analysis on Incremental Classifier.} We further investigated our proposed Incremental Classifier, IC, regarding the average thresholds which are the criteria for increasing nodes for each label, and the number of increased nodes in Figure \ref{fig:thre}.
% and the effect of dynamic threshold as shown in Table \ref{tab:dt}. 
In Figure \ref{fig:thre}, the ratio of increased nodes, (\ie the number of increased nodes relative to the total number of classes in the entire tasks), was the most numerous in DomainNet, indicating that it is composed of images with huge domain gap among domains. In CORe50 and iDigits which have relatively smaller domain gaps, the nodes were increased less since existing nodes in the classifier can accommodate knowledge of multiple domains when a new task arrives. Also, the average thresholds of classes in the entire task are demonstrated in Figure \ref{fig:thre}. Since domain differences are big in the order of DomainNet, iDigits, and CORe50, the average thresholds follow the same order. The higher the threshold, the more the model increases the nodes because the number of classes that do not exceed the threshold increase.

\subsubsection{Ablation of Dynamic Threshold.} The impact of using dynamic threshold (DT) in IC is demonstrated in Table \ref{tab:dt}. The comparison was conducted in the setting in which the threshold value is set to a constant of 0.5. The strategy of setting threshold dynamically was effective in leveraging IC, by successfully deciding the optimal threshold value on the basis of domain differences calculated using accuracies per class.
% \begin{figure}[t]
% \centering
% \includegraphics[width=\columnwidth]{figure/nodes.png}
% \caption{Results of routing nodes of DomainNet in VIL.}
% \label{fig:nodes}
% \end{figure}

\subsubsection{Incremental Classifier Compared to Standard CIL Strategy.} The mechanism of Incremental Classifier (IC) is quite different from a standard strategy in CIL research. The existing strategy in CIL only increases output nodes corresponding to new classes naively when new tasks arrive.
However, our proposed IC can increase output nodes of both old and new classes to prevent semantic drift of the classifier while learning diverse domains, with our novel strategies that decide when to expand each node and handle the node selection issue for each class.
In this way, IC can tackle the challenge successfully while learning various domains associated with a single class, with dynamic thresholding and knowledge distillation to mitigate catastrophic forgetting effectively. 
Consequently, IC successfully tackles forgetting in the classifier caused by various domains while the existing strategy in CIL cannot.

\begin{figure*}[!h]
\centering
\includegraphics[width=0.9\textwidth]{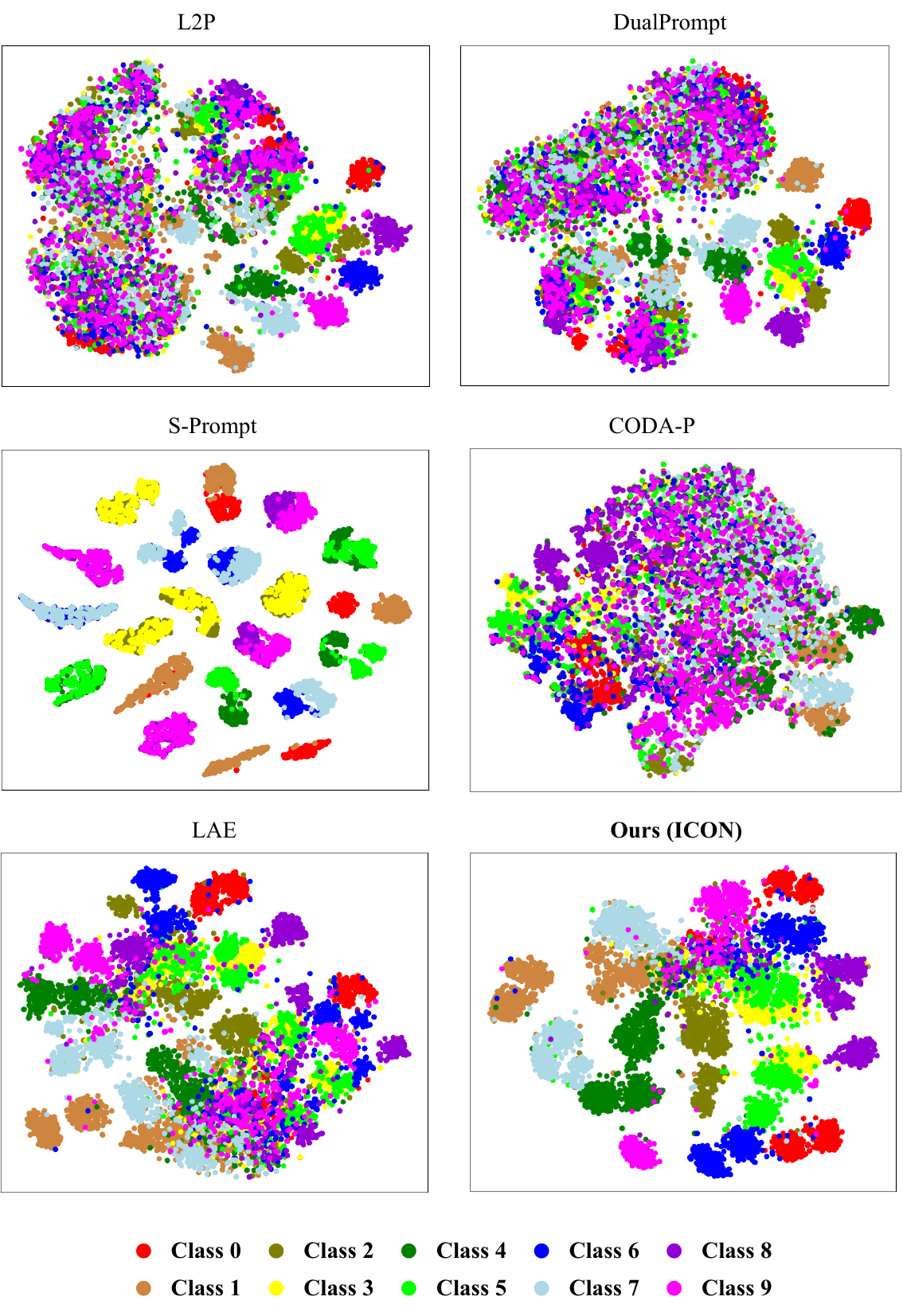}
\caption{t-SNE visualization on the resulting feature spaces of L2P \cite{wang2022learning}, DualPrompt \cite{wang2022dualprompt}, S-Prompts \cite{wang2022sprompts}, CODA-P \cite{smith2023coda}, LAE \cite{gao2023unified} and proposed ICON on the iDigits dataset in VIL scenario.}
\label{fig:tsne}
\end{figure*}

\begin{table*}[!t]
\caption{Additional experiments on CNN.}
\renewcommand{\arraystretch}{1.2}
\renewcommand{\tabcolsep}{1.0}
\resizebox{\textwidth}{!}{%
\begin{tabular}{lcccccccccc}
\specialrule{.1em}{.1em}{.1em} 
\multicolumn{1}{l}{\multirow[b]{1.8}{*}{Dataset}} & \multicolumn{2}{c}{Fine-tuning} & \multicolumn{2}{c}{EWC \cite{kirkpatrick2017overcoming}} & \multicolumn{2}{c}{LwF \cite{li2016learning}} & \multicolumn{2}{c}{LAE \cite{gao2023unified}} &  \multicolumn{2}{c}{\textbf{ICON (Ours})} \\ 
\cmidrule(lr){2-3} \cmidrule(lr){4-5} \cmidrule(lr){6-7} \cmidrule(lr){8-9} \cmidrule(lr){10-11} 
\multicolumn{1}{c}{} & Avg. Acc↑ & Forgetting↓ & Avg. Acc↑ & Forgetting↓ & Avg. Acc↑ & Forgetting↓ & Avg. Acc↑ & Forgetting↓ & Avg. Acc↑ & Forgetting↓ \\ 
\specialrule{.05em}{.1em}{.1em} 
iDigits \cite{volpi2021continual}& 23.22$\pm$0.24 & 25.13$\pm$0.94 & 29.91$\pm$0.64 & 16.63$\pm$0.98 & 28.64$\pm$0.43 & 17.94$\pm$0.51 & 30.43$\pm$0.36 & 15.62$\pm$1.12 & \textbf{37.34$\pm$0.73} & \textbf{7.55$\pm$0.47} \\
CORe50 \cite{lomonaco2017core50}& 13.49$\pm$0.69 & 27.61$\pm$0.08 & 14.14$\pm$1.34 & 23.98$\pm$0.36 & 23.29$\pm$0.81 & 12.22$\pm$0.61 & 37.36$\pm$0.49 & 6.34$\pm$0.55 & \textbf{45.67$\pm$0.66} & \textbf{5.92$\pm$0.41} \\
DomainNet \cite{peng2019moment}& 11.34$\pm$0.37 & 20.91$\pm$0.42 & 13.74$\pm$0.73 & 13.83$\pm$0.89 & 14.83$\pm$1.53 & 13.16$\pm$0.36 & 17.68$\pm$0.67 & 12.46$\pm$0.69 & \textbf{23.73$\pm$0.55} & \textbf{7.32$\pm$0.28} \\
\specialrule{.1em}{.1em}{.1em} 
\end{tabular}%
}
\label{tab:cnn}
\end{table*}

\subsubsection{Architecture Generalizability.} Existing prompt-based methods are not flexible enough to be combined with architectures other than the Transformer family. On the other hand, our proposed ICON uses a bottleneck adapter that is sufficiently applicable to CNN and others. Therefore, we applied ICON to CNN structure without pre-trained weights other than ViT and demonstrated its performance with Fine-tuning, EWC \cite{kirkpatrick2017overcoming}, LwF \cite{li2016learning} and LAE \cite{gao2023unified} similarly applicable in the CNN structure. We used ResNet-152 (60M) \cite{he2016deep}, which has a similar number of parameters as ViT-B/16 (86M). The bottleneck adapter was implemented with 1$\times$1 convolution layers for up and down projection and inserted in the shallower 23 of 50 convolution blocks parallelly. As shown in Table \ref{tab:cnn}, our proposed ICON also achieved significantly better performance compared to other IL methods based on CNN architecture, emphasizing the broader applicability of ICON beyond being limited to ViT.

\begin{table}[!t]
\renewcommand{\arraystretch}{.8}
\renewcommand{\tabcolsep}{1.5mm}
\centering
\caption{Average accuracy on various number of shifts per task.}
    \resizebox{0.6\textwidth}{!}{% 
        \begin{tabular}{cccc}
        \specialrule{.1em}{.1em}{.1em} 
        \multirow{2}{*}{\begin{tabular}[c]{@{}c@{}}\# shifts \\ per task\end{tabular}} & \multirow{2}{*}{iDigits \cite{volpi2021continual}} & \multirow{2}{*}{CORe50 \cite{lomonaco2017core50}} & \multirow{2}{*}{DomainNet \cite{peng2019moment}} \\
        &&&\\
        \specialrule{.05em}{.1em}{.1em} 
         1 &\textbf{75.11}&\textbf{83.18}&53.37\\
         2 &71.77&82.66&53.23 \\
         4 &69.24&82.70&\textbf{53.89} \\
         6 &71.82&82.28&53.66\\
         8 &69.06&82.51&53.68 \\
        10 &68.45&81.70&53.61 \\
        \specialrule{.1em}{.1em}{.1em} 
        \end{tabular} %
    }
    \label{tab:shift}
\end{table}

\subsubsection{Multiple Shifts per Task.} We further explored using shifts more than one that are saved in the shift pool for each task, as shown in Table \ref{tab:shift}. For iDigits, when used more than one shift for a task, the performance drops showing that having too many shifts in the shift pool can cause too strict regularization in CAST, and CORe50 has similar results. For DomainNet, the performance was robust against the number of shifts saved per task.

\subsubsection{Representation Spaces of Competing Methods.}
As we mentioned in the main paper, the model faces intra-class domain confusion and inter-domain confusion in the proposed VIL scenario, and these confusions are very likely to appear in a mixed state in feature representation space. Therefore, we visualized the feature representation spaces of competing methods in the VIL scenario using t-SNE (Figure \ref{fig:tsne}). As shown in Figure \ref{fig:tsne}, in the case of prompt-based methods, such as L2P \cite{wang2022learning} and DualPrompt \cite{wang2022dualprompt}, only the most recently learned classes are biased and separated. However, unlike other prompt-based methods, CODA-P \cite{smith2023coda} shows a lot of mixtures, which we expect to be the result of the weighted sum of prompt using attention. S-Prompts \cite{wang2022sprompts} strongly rely on task-specific prompt and selection mechanisms, therefore, representations of classes learning in the same task are overlapped. LAE \cite{gao2023unified} using an adapter, shows relatively more separation compared to the aforementioned methods, but there are still many mixed representations. In contrast, our proposed ICON produces much more separated class-specific representation subspaces. This demonstrates that ICON has well-accumulated knowledge without interfering with previously learned knowledge away from intra-class domain confusions and inter-domain confusions.

\section{Additional Ablation Results}\label{sec:ablation}

\begin{table}[!t]
    \captionof{table}{Average accuracy on various locations of adapter.}
    \centering
    \renewcommand{\arraystretch}{1.2}
    \renewcommand{\tabcolsep}{6mm}
    \resizebox{0.9\columnwidth}{!}{% 
    \begin{tabular}{lccc}
        \specialrule{.1em}{.1em}{.1em} 
        \multirow{1}{*}{Location}  & \multicolumn{1}{c}{iDigits \cite{volpi2021continual}}  & \multicolumn{1}{c}{CORe50 \cite{lomonaco2017core50}} & \multicolumn{1}{c}{DomainNet \cite{peng2019moment}}\\
        \specialrule{.05em}{.1em}{.1em} 
        First 5 (\textbf{Ours}) & \textbf{75.11} & \textbf{83.18} & \textbf{53.37} \\
        Last 5& 52.94 & 77.79  & 50.31 \\
        All & 62.57& 83.10 & 55.14\\
        \specialrule{.1em}{.1em}{.1em} 
    \end{tabular} %
    }
    \label{tab:location}
\end{table}

\subsubsection{Results on Various Locations of Adapter.} We conducted experiments of ICON with different locations of adapters (Table \ref{tab:location}). For all datasets, inserting adapters to the first 5 layers of the backbone was the best, and the performances significantly dropped when used them in the last 5 layers of the backbone. The result indicates that adapting to new task is optimal in the early layers, and using them in the later layers prevents flexible adjustment of representations to each new task. 
% Also, when used adapters to the whole layers of the backbone, the performance dropped significantly. This result implies that it is more effective to use adapters in the early layers by restraining the range of drift in feature space during training.

\subsubsection{Hyperparameter Sensitivity.} 
In this section, we demonstrated the result of our experiments with different hyperparameter values, $\alpha$, $\beta$, and $\gamma$ which are used in the loss function. The hyperparameter $\alpha$ adjusts the impact of knowledge distillation from the previous classifier which is involved in our proposed Incremental Classifier. 

As shown in Table \ref{tab:alpha}, using small $\alpha$ prevents fully leveraging the benefit of knowledge distillation, while the excessive effect of knowledge distillation learning the current task. We set $\alpha = 1.00$ for all datasets as default. We also explored the impact of CAST loss via the coefficient $\beta$ that controls the power of regularization of the direction of the current task in Table \ref{tab:beta}. For iDigits and CORe50, the performance was the best when $\beta = 0.05$, while the performance of DomainNet was the best when $\beta = 0.01$, indicating relatively the weak intensity of regularization allows learning more difficult tasks (DomainNet). Lastly, the impact of the scaling factor $\gamma$ used when deciding the threshold in IC is demonstrated in Table \ref{tab:gamma}. Using too small $\gamma$ can cause overwriting of the classifier by not increasing nodes in the classifier dynamically even when the domain difference is huge. Meanwhile, using too big $\gamma$ can degrade the performance since the model can be confused while selecting which logit to use for a single label at inference time, after increasing nodes too easily even when not necessary.

\begin{table}[!t]
    \captionof{table}{Average accuracy on various $\alpha$.}
    \centering
    \renewcommand{\tabcolsep}{8mm}
    \resizebox{0.9\columnwidth}{!}{% 
    \begin{tabular}{cccc}
    \specialrule{.1em}{.1em}{.1em} 
    \multirow{1}{*}{$\alpha$}  & \multicolumn{1}{c}{  iDigits \cite{volpi2021continual}  }  & \multicolumn{1}{c}{  CORe50 \cite{lomonaco2017core50}  } & \multicolumn{1}{c}{DomainNet \cite{peng2019moment}}\\
    \specialrule{.05em}{.1em}{.1em} 
      0.25  &74.73&81.30&52.45 \\
      0.50  &74.90&82.51&53.21 \\
      1.00  &\textbf{75.11}&83.18&\textbf{53.37} \\
      1.50  &74.97&\textbf{83.23}&53.06\\
      2.00  &74.88&82.89&52.80 \\
    \specialrule{.1em}{.1em}{.1em} 
    \end{tabular} %
    }
    \label{tab:alpha}
\end{table}

% \begin{minipage}[t]{.3\textwidth}
\begin{table}[!t]
    \captionof{table}{Average accuracy on various $\beta$.}
    \centering
    \renewcommand{\tabcolsep}{7mm}
    \resizebox{0.9\columnwidth}{!}{% 
    \begin{tabular}{cccc}
    \specialrule{.1em}{.1em}{.1em} 
    \multirow{1}{*}{$\beta$}  & \multicolumn{1}{c}{  iDigits \cite{volpi2021continual}  }  & \multicolumn{1}{c}{  CORe50 \cite{lomonaco2017core50}  } & \multicolumn{1}{c}{DomainNet \cite{peng2019moment}}\\
    \specialrule{.05em}{.1em}{.1em} 
      0.01  &74.51&79,85&\textbf{53.37} \\
      0.02  &74.78&80.55&53.21 \\
      0.03  &74.22&82.15&53.17 \\
      0.04  &73.95&83.08&53.19 \\
      0.05  &\textbf{75.11}&83.18& 53.27\\
      0.06  &74.69&83.21&53.07 \\
      0.07  &74.95&\textbf{83.50}&53.03 \\
      0.08  &73.89&83.02&52.75 \\
      0.09  &74.06&83.24&52.90\\
      0.10  &74.32&83.11&52.66 \\
    \specialrule{.1em}{.1em}{.1em} 
    \end{tabular} %
    }
    \label{tab:beta}
\end{table}

\begin{table}[!t]
    \captionof{table}{Average accuracy on various $\gamma$.}
    \centering
    \renewcommand{\tabcolsep}{7mm}
    \resizebox{0.9\columnwidth}{!}{% 
    \begin{tabular}{cccc}
    \specialrule{.1em}{.1em}{.1em} 
    \multirow{1}{*}{$\gamma$}  & \multicolumn{1}{c}{  iDigits \cite{volpi2021continual}  }  & \multicolumn{1}{c}{  CORe50 \cite{lomonaco2017core50}  } & \multicolumn{1}{c}{DomainNet \cite{peng2019moment}}\\
    \specialrule{.05em}{.1em}{.1em} 
      0.5  &72.00&80.74&\textbf{53.53}\\
      1.0  &73.12&81.23&53.41 \\
      1.5  &74.51&82.70&53.22 \\
      2.0  &\textbf{75.11}&\textbf{83.18}&53.37\\
      2.5  &75.08&82.52&53.12 \\
      3.0  &74.99&82.74&52.99 \\
    \specialrule{.1em}{.1em}{.1em} 
    \end{tabular} %
    }
    \label{tab:gamma}
\end{table}

\section{Limitations and Future Work}\label{sec:limitations} Despite achieving noticeable performance and successfully resolving the problem from the absence of prior knowledge of the following tasks, our work has a few limitations as well. We conducted experiments on widely used three benchmarks, that can be configured VIL scenario. Then we tried to experiment with additional datasets, but other widely used datasets have the following problems when constructing the VIL scenario. ImageNet-R \cite{hendrycks2021many} and VLCS \cite{torralba2011unbiased} have severe imbalance problems, and thus in some classes, it becomes a few-shot (about 1 to 5) learning task when constructing a VIL scenario. Moreover, PACS \cite{li2017deeper} and OfficeHome \cite{venkateswara2017deep} have an insufficient number of classes (both have 7 classes) to strictly evaluate incremental scenarios and have imbalance problems too. Thus, the VIL scenario requires more complex and well-organized large benchmarks to evaluate the effectiveness of VIL methods and encourage the advances of this real-world challenge. 

Furthermore, in IC, there can be other algorithms to replace max pooling which are more effective based on further analysis even though they work well currently. In the same way, deciding the number of clusters in CAST as the sequential tasks increase can be addressed in the future work. Also, while only a single domain and a single group of classes increase in the current VIL setting, having more than a single domain and a single group of classes can be included in the future work as well.

\section{Hyperparameters of Comparison Models.}\label{sec:hyperparam_comparison}
\begin{enumerate}
    \item L2P \cite{wang2022learning}
        \begin{itemize}[label={$\bullet$}]
            \item iDigits
                \begin{itemize}[label={-}]
                    \item Prompt pool size: 20
                \end{itemize}
            \item CORe50
                \begin{itemize}[label={-}]
                    \item Prompt pool size: 40
                \end{itemize}
            \item DomainNet
                \begin{itemize}[label={-}]
                    \item Prompt pool size: 30
                \end{itemize}
            \item Common
                \begin{itemize}[label={-}]
                    \item Prompt top-$K$: 5
                    \item Prompt length: 5
                    \item Batch size: 16
                \end{itemize}
        \end{itemize}
    \bigskip
    \item S-Prompts \cite{wang2022sprompts}
        \begin{itemize}[label={$\bullet$}]
            \item iDigits
                \begin{itemize}[label={-}]
                    \item Number of clusters: 4
                \end{itemize}
            \item CORe50
                \begin{itemize}[label={-}]
                    \item Number of clusters: 8
                \end{itemize}
            \item DomainNet
                \begin{itemize}[label={-}]
                    \item Number of clusters: 6
                \end{itemize}
            \item Common
                \begin{itemize}[label={-}]
                    \item Prompt length: 10
                    \item Batch size: 128
                \end{itemize}
        \end{itemize}
    \bigskip
    \item DualPrompt \cite{wang2022dualprompt}
        \begin{itemize}[label={$\bullet$}]
            \item iDigits
                \begin{itemize}[label={-}]
                    \item E-Prompt pool size: 20
                \end{itemize}
            \item CORe50
                \begin{itemize}[label={-}]
                    \item E-Prompt pool size: 40
                \end{itemize}
            \item DomainNet
                \begin{itemize}[label={-}]
                    \item E-Prompt pool size: 30
                \end{itemize}
            \item Common
                \begin{itemize}[label={-}]
                    \item G-Prompt layer index: [0, 1]
                    \item G-Prompt length: 5
                    \item E-Prompt layer index: [2, 3, 4]
                    \item E-Prompt length: 5
                    \item E-Prompt top-$K$: 1
                    \item Batch size: 24
                \end{itemize}
        \end{itemize}
    \bigskip
    \item CODA-P \cite{smith2023coda}
        \begin{itemize}[label={$\bullet$}]
            \item iDigits
                \begin{itemize}[label={-}]
                    \item E-Prompt pool size: 20
                \end{itemize}
            \item CORe50
                \begin{itemize}[label={-}]
                    \item E-Prompt pool size: 40
                \end{itemize}
            \item DomainNet
                \begin{itemize}[label={-}]
                    \item E-Prompt pool size: 30
                \end{itemize}
            \item Common
                \begin{itemize}[label={-}]
                    \item G-Prompt length: 0
                    \item E-Prompt length: 8
                    \item Batch size: 128
                \end{itemize}

        \end{itemize}
    \bigskip
    \item LAE \cite{gao2023unified}
        \begin{itemize}[label={$\bullet$}]
            \item Common
                \begin{itemize}[label={-}]
                    \item Adapter location: [0, 1, 2, 3, 4] of Multi-head Self-Attention layer
                    \item Adapter downsample: 5
                    \item EMA decay: 0.9999
                    \item Batch size: 24
                \end{itemize}
        \end{itemize}
\end{enumerate}

{
    % \small
    \bibliographystyle{splncs04}
    % \bibliography{main}

}

\par\vfill\par